\documentclass[journal]{IEEEtran}

\usepackage[utf8]{inputenc}
\usepackage[T1]{fontenc}
\usepackage{siunitx}
\usepackage{graphicx}
\usepackage[colorlinks=true, allcolors=blue]{hyperref}
\usepackage{amsmath}
\usepackage{amsfonts}
\usepackage{amssymb}
\usepackage{gensymb}
\usepackage{booktabs}
\usepackage{soul} 
\usepackage{subcaption}
\usepackage{stfloats}
\usepackage{booktabs}
\usepackage{enumitem}
\usepackage{pifont}
\usepackage{tikz,xcolor}
\usepackage{makecell}

\graphicspath{{graphics/}}

\definecolor{lime}{HTML}{A6CE39}
\DeclareRobustCommand{\orcidicon}
{
    \begin{tikzpicture}
    \draw[lime, fill=lime] (0,0) circle [radius=0.16] 
    node[white] {{\fontfamily{qag}\selectfont \tiny ID}};    \draw[white, fill=white] (-0.0625,0.095) circle [radius=0.007];    
    \end{tikzpicture}
    \hspace{0mm}}
\foreach \x in {A, ..., Z}{%
    \expandafter\xdef\csname orcid\x\endcsname{\noexpand\href{https://orcid.org/\csname orcidauthor\x\endcsname}{\noexpand\orcidicon}}
}

\setlist[enumerate]{itemsep = 0pt, parsep = 0pt, topsep = 0pt} 
\setlist[itemize]{itemsep = 0pt, parsep = 0pt, topsep = 0pt} 

\begin{document}

\title{Automated Parking Planning with Vision-Based BEV Approach}

\author{Yuxuan Zhao}

\markboth{Journal of \LaTeX\ Class Files,~Vol.
}%
{Shell \MakeLowercase{\textit{et al.}}: Bare Demo of IEEEtran.cls for IEEE Journals}

\maketitle

\begin{abstract}

Automated Valet Parking (AVP) is a crucial component of advanced autonomous driving systems, focusing on the endpoint task within the "human-vehicle interaction" process to tackle the challenges of the "last mile".
The perception module of the automated parking algorithm has evolved from local perception using ultrasonic radar and global scenario precise map matching for localization to a high-level map-free Birds Eye View (BEV) perception solution.
The BEV scene places higher demands on the real-time performance and safety of automated parking planning tasks.

This paper proposes an improved automated parking algorithm based on the A* algorithm, integrating vehicle kinematic models, heuristic function optimization, bidirectional search, and Bezier curve optimization to enhance the computational speed and real-time capabilities of the planning algorithm.
Numerical optimization methods are employed to generate the final parking trajectory, ensuring the safety of the parking path. 
The proposed approach is experimentally validated in the commonly used industrial CARLA-ROS joint simulation environment.

Compared to traditional algorithms, this approach demonstrates reduced computation time with more challenging collision-risk test cases and improved performance in comfort metrics.
\end{abstract}


\IEEEpeerreviewmaketitle

\section{Introduction}

The automotive industry has become one of the engines of economic growth and a major driving force, playing a key role in industrial upgrading and national competitiveness \cite{lan2023virtual,yi2024key}.
The vigorous development of the automotive industry is conducive to providing high-quality employment opportunities, increasing people's income, promoting social development, enhancing manufacturing strength, and boosting national comprehensive competitiveness.
Autonomous driving is the current trend in the automotive industry and has become a hot topic in global innovation. 
Currently, both industry and academia primarily divide autonomous driving systems into three modules: perception mapping, path planning, and tracking control \cite{lan2023end}.
The perception module is responsible for using sensor combinations to enable autonomous vehicles to perceive obstacles, traffic lights, traffic signs, and other information in the surrounding environment. 
The planning module is responsible for completing the vehicle's driving path planning based on the perceived information and some a priori data, such as high-precision maps.
The control module is responsible for providing instructions to the hardware to ensure that the vehicle correctly follows the predetermined trajectory.

In recent years, many domestic and international technology companies such as Huawei, Tesla, Google, Baidu, DJI, and XPeng have entered the field of autonomous driving, with numerous universities and research institutes dedicating significant efforts to intelligent driving research.
In 2020, the National Development and Reform Commission and the Ministry of Industry and Information Technology issued the \textit{Smart Car Innovation Development Strategy} \cite{ref1} to guide the future direction of intelligent automobile development. 
By this plan, mass production of autonomous intelligent vehicles should be realized, and highly automated intelligent vehicles should achieve market applications in specific environments.

Automated parking is an important and challenging function in autonomous driving systems.
Automated parking systems can greatly simplify the parking process, optimize the driving experience for consumers, and significantly reduce the probability of traffic accidents during parking, providing a safer and more convenient experience. 
Inexperienced drivers, in particular, can benefit greatly from automated parking systems.
The widespread adoption of automated parking systems can promote the popularization of small, efficient unmanned automatic parking lots, thereby saving precious land resources. Up to now, solutions specifically designed for automated parking tasks are not widespread, "combined driving and parking" concepts are still in the research and development stage, and companies and vehicle models with automated parking as the main selling point are rare in the industry.
Therefore, automated parking is an area that requires and deserves more research dedication.

In traditional methods, autonomous driving systems usually rely on high-precision maps for perceiving the surrounding static environment. 
High-precision maps have the characteristics of high accuracy, richness, and real-time performance, providing abundant precise road information and serving as a geographical reference for autonomous driving tasks \cite{ref6}.
These high-precision maps offer:

\begin{enumerate}
    \item Perception: High-precision maps provide precise road and obstacle information, giving vehicles the ability to anticipate beyond sensor limitations, acquire information beyond line of sight, and compensate for the upper limits of perception capabilities \cite{ref7}.
    \item Positioning: Autonomous driving vehicles can compare the information obtained by on-board sensors with the corresponding records in high-precision maps at any time to greatly improve the positioning accuracy of autonomous driving vehicles, providing precise information for subsequent downstream tasks.
\end{enumerate}

However, removing high-precision maps has become a trend in the autonomous driving industry as a response to the inherent drawbacks of high-precision maps:

\begin{enumerate}
    \item Cost issues: Maintaining high-precision maps requires a fleet of map acquisition vehicles with individual costs exceeding one million RMB, constantly collecting and updating data for the country's continuously changing roads.
    The overall fleet expenses can easily exceed billion-level amounts, making it nearly impossible to obtain and maintain high-precision maps containing all or most of road data.
    \item Qualifications for map surveying: For national security reasons, the government has strict controls on high-precision map surveying rights. 
    Complex audit and verification mechanisms restrict the quick updates and release of high-precision maps.
\end{enumerate}

In the context of the development trend without high-precision maps and without lidar technology, the shortcomings of traditional methods for automated parking, such as low real-time performance and insufficient planning safety, stand out, posing significant challenges to large-scale promotions.
These urgent problems greatly hinder the automated parking module from catching up with the trend without high-precision maps, slowing down the process of making automated parking cost-effective and popular.

To address the poor perception performance in underground parking scenarios and the lack of precise results in traditional planning methods, this paper introduces the industry-leading Bird's Eye View (BEV) \cite{ref9} perception method and optimizes it for the characteristics of parking environments based on traditional parking planning algorithms.

This paper proposes an automatic parking planning method based on a purely visual BEV local map.
The method uses a BEV + Transformer \cite{ref10} + Occupancy \cite{ref11} perception solution for mapping operations. The paper presents an optimized planning algorithm based on the A* algorithm to carry out automatic parking navigation path planning tasks, aiming to reduce algorithm runtime and improve the real-time capabilities of the A* algorithm.
Additionally, numerical optimization methods are employed to address the high-precision requirements of parking tasks.
The proposed method can handle reverse parking task scenarios that are challenging for traditional perception and planning methods, providing targeted solutions to the existing issues.

\section{Related Work}
\label{sec:related_work}

\subsection{Characteristics of Automated Parking}

The subdivision tasks of the automated parking module in autonomous driving have their own uniqueness.
By 2024, automated parking has become a hot topic in the field of autonomous driving, and an important function that major companies in the industry are actively researching and implementing. 
Automated parking tasks extensively involve perception and planning in garage or parking lot scenarios. Common parking lot scenarios exhibit the following characteristics: poor lighting, obstruction of visibility by obstacles such as pillars \cite{ref12}. 
In such scenarios, traditional perception methods are inevitably subject to interference, leading to suboptimal perception results.
Due to cost considerations, in real-world scenarios, automated parking systems must address the challenge of lacking high-precision maps for assisting in perception and planning.
However, traditional perception methods without the assistance of high-precision maps often result in fragmented camera data, low accuracy, and an inability to meet the subsequent mapping and planning requirements.
Therefore, this paper aims to use locally generated maps to achieve the basic functions of high-precision maps \cite{xu2019online}.

Parking scenarios are widely considered to be highly challenging driving tasks, where vehicles must complete a series of actions including but not limited to backing up and precise maneuvering within narrow spaces, leading to a high probability of collisions \cite{ref14}. 
The complexity lies in planning the appropriate trajectory to park in the parking space and finely controlling the vehicle's movement in tight spaces to avoid obstacles such as pillars, other vehicles, pedestrians, etc.
In traditional methods, automated parking tasks require the use of real-time updated online maps or pre-downloaded high-precision offline maps to assist the automated parking system in completing path planning and self-control.

Automated parking scenarios also pose demands on planning algorithms. 
There are many uncontrollable factors during parking, such as issues with the steering system's speed and accuracy, and changes in the position of reference obstacles, leading to deviations from the planned parking trajectory.
Hence, this paper requires the planning algorithm during parking to dynamically correct trajectories in real-time and even perform re-planning \cite{ref15}. Current parking scenarios mainly consist of perpendicular parking and parallel parking, and the planning algorithms need to support both of these major parking scenarios simultaneously.
Moreover, the narrow spatial environment of underground parking garages can easily cause collisions if planning accuracy is not high enough.

The automated parking perception module needs to use more precise bounding boxes to approximate the shapes of the vehicle and obstacles, and accurately locate the actual position of the vehicle in the underground garage to avoid collisions. 
The planning part needs to find the trajectory in the cluttered environment of the garage. Hence, it is necessary to meet multiple requirements for perception, positioning, motion planning, and actuator precision to further advance automated parking technology.
Due to these reasons, the cost of automated parking functionality is relatively high and cannot be widely deployed on a large scale \cite{ref15}.

In summary, the automated parking tasks in the local map have two major characteristics:

\begin{enumerate}
    \item Local map perception requires planning algorithms to have high real-time performance and fast computation speed.
    \item Automated parking tasks require planning algorithms to generate high-quality trajectories for stable and safe driving.
\end{enumerate}

The automated parking solution proposed in this paper must consider the inherent complexity of automated parking tasks and the high requirements for positioning and planning accuracy.
The pure visual BEV perception method proposed in this study can provide sufficiently accurate perception information, achieving high-precision perception and positioning in the absence of high-precision maps. 
Additionally, the improved A* algorithm proposed for generating parking navigation paths has high real-time performance, and numerical optimization methods can generate high safety and comfort parking trajectories, meeting the precision requirements of automated parking tasks.

\subsubsection{Traditional Planning Algorithms}
\label{sec:Traditional_Planning_Algorithms}

In the context of autonomous driving, the planning problem requires the vehicle to start from an initial position and reach a target position while satisfying given global and local constraints \cite{ref17}.
In autonomous driving, occupancy grids or grid maps are often used to represent grid maps based on the vehicle’s coordinate system.
Under this premise, various algorithms have been applied in the field of autonomous driving.
Broadly classified, they can be divided into graph search algorithms, random sampling algorithms, interpolating curve planning algorithms, and numerical optimization algorithms \cite{ref18}.

Graph search algorithms construct a path from the start point to the endpoint based on a known environmental map and obstacle information within the map. 
Common graph search algorithms include the Dijkstra algorithm, A* algorithm, and state lattice planning.

The Dijkstra algorithm is a graph search algorithm for finding the shortest path from a single source in a graph.
The Dijkstra algorithm starts from the initial point and performs a breadth-first search, checking the vehicle and surrounding grids. 
This algorithm can search for the shortest path between any two points in the graph \cite{ref19} and guarantees the generation of an optimal path from the vehicle’s position to the planned endpoint, provided a feasible path exists. 
This algorithm has been implemented in the Darpa Urban Challenge \cite{ref20}.
If the autonomous vehicle has a pre-obtained static semantic map containing road information, the Dijkstra algorithm can ensure the generation of the optimal planned path.

The A* algorithm is a heuristic graph search algorithm improved from the Dijkstra algorithm.
It defines the weight of nodes and can achieve faster searches.
This is a widely used planning algorithm that performs well when global map information is available.
However, as the map size increases, the search cost of the A* algorithm grows exponentially, performing poorly in large-scale planning problems.

State lattice planning discretizes the continuous space into a state lattice map, simplifying the complexity of planning in real space.
In the state lattice, the connections between each vertex are generated based on the vehicle’s kinematic model, ensuring that all paths are feasible \cite{ref21}.
This avoids the issue of some planning methods generating paths that cannot be executed by the control module. 
It adds kinematic constraints on top of ordinary grid maps, ensuring state continuity.

Random sampling algorithms search for feasible paths by randomly sampling paths in the space and iterating continuously. 
Rapidly-exploring Random Tree (RRT) and Probabilistic Road Map (PRM) algorithms are two of the most common random sampling algorithms.

The Rapidly-exploring Random Tree (RRT) algorithm is a sampling-based planning algorithm. It conducts random searches in non-convex high-dimensional spaces, using the vehicle’s position as the root node and selecting reachable points as new leaf nodes to generate a random expansion tree, allowing the search tree to grow towards the navigation endpoint to obtain a feasible path. 
It enables fast planning in semi-structured spaces, with fast computation speed and the ability to asymptotically converge to the optimal solution \cite{ref24}. 
Since the vehicle in the real environment may not be able to execute the movements required by the random search tree, in practice, constraints such as steering angle are often added manually, and interpolation methods are used to fit the initially generated path to make the final path smoother \cite{ref25}.

The Probabilistic Road Map (PRM) algorithm is another sampling-based algorithm. This algorithm randomly generates a large number of nodes in the search space and then connects all the nodes. 
It then selects a collision-free path from the start point to the endpoint in the connected graph to generate a feasible path. 
However, the drawback of this algorithm is that collision detection takes a long time when there are too many obstacles \cite{ref26}.

Interpolation planning algorithms include various methods such as circular-arc and straight-line method, clothoid planning, polynomial planning, and spline planning.
Interpolating curve planning requires a pre-defined set of nodes, then uses interpolation algorithms to generate a series of navigation trajectories that satisfy trajectory continuity and vehicle kinematic constraints.
Interpolation planning methods achieve the technology of generating smooth curved paths while ensuring continuous and smooth movement in terms of speed and acceleration \cite{ref27}, making them suitable for scenarios with requirements for obstacle avoidance, path quality, and trajectory comfort. 
Using geometric curves for parking path planning is relatively simple to implement but is ideal for standard parking spaces, with limited effectiveness for irregular spaces, spaces blocked by numerous obstacles, and narrow spaces \cite{ref28}.

The two-segment circular-arc method and the circular-arc and straight-line combined method are simple parallel parking path planning methods. 
The main idea is to use two segments of circular arcs as the basic path and then directly connect the two arcs or use a straight path to connect them.
This algorithm is prone to curvature discontinuities, causing the planned path to have discontinuous curvature, making it difficult for the control module to satisfy kinematic constraints and making it challenging for real-world vehicles to follow the path generated by this algorithm. 
Therefore, when using this method for path planning, it is necessary to combine other interpolation curve planning methods to fit the points of curvature discontinuity and solve the issue of curvature discontinuity \cite{ref30}.

Clothoid planning uses clothoid curves defined by Fresnel integrals for path planning \cite{ref31}. 
These curves can define trajectories with linearly varying curvature, smoothly transitioning between curved and straight sections of the trajectory. 
This method is easy to implement and ensures the path has excellent approximation accuracy.

\begin{table*}[!ht] \small \centering
\setlength\tabcolsep{2pt} \renewcommand{\arraystretch}{1.0}
    \caption{Performance Comparison of Traditional Algorithms and the Proposed Method. \checkmark indicates the algorithm possesses the characteristic, \ding{53} indicates it does not, ? indicates the algorithm possesses the characteristic in some cases.}
    \begin{tabular}{lccccc} \toprule
        Algorithm & Optimal Path & Fast Computation & Continuous Path & Kinematic Feasibility & No Pre-training  \\ \midrule
        Dijkstra \cite{ref19,ref20} & \checkmark & \ding{53} & \ding{53} & \ding{53} & \checkmark  \\ 
        A* & \checkmark & ? & \ding{53} & \ding{53} & \checkmark  \\ 
        State Lattice Planning \cite{ref21} & \checkmark & \ding{53} & \checkmark & \checkmark & \checkmark  \\ 
        RRT \cite{ref24,ref25} & ? & \checkmark & \ding{53} & \ding{53} & \checkmark  \\ 
        PRM \cite{ref26} & ? & ? & \ding{53} & \ding{53} & \checkmark  \\ 
        Arc-Line \cite{ref30} & \ding{53} & \checkmark & \checkmark & ~ & \checkmark  \\ 
        Clothoid Planning \cite{ref31} & ~ & ~ & \checkmark & \checkmark & \checkmark  \\ 
        Polynomial Planning \cite{ref32} & ~ & \checkmark & \checkmark & ~ & \checkmark  \\ 
        Spline Planning \cite{ref33,ref36} & ~ & \checkmark & \checkmark & ~ & \checkmark  \\ 
        Numerical Optimization & \checkmark & \ding{53} & \checkmark & \checkmark & \checkmark  \\ 
        Artificial Intelligence \cite{ref39} & \checkmark & ~ & \checkmark & \checkmark & \ding{53}  \\ 
        \textbf{Proposed} & \checkmark & \checkmark & \checkmark & \checkmark & \checkmark  \\ \bottomrule
    \end{tabular}
    \label{tab:plan_methods}
\end{table*}

Polynomial planning satisfies the constraints given by the interpolation points. 
This type of method uses polynomials of different degrees to achieve vertical constraints, horizontal constraints, and generate safe driving trajectories \cite{ref32}. 
It performs well in fitting position, angle, and curvature constraints.

Spline planning uses piecewise polynomial parameters to generate a set of high-order continuously differentiable polynomials by linearly combining spline basis functions, such as cubic spline curves and B-spline curves \cite{ref33}. 
The connection points of spline segments have high smoothness. Currently, spline planning has been applied to autonomous driving projects in various scenarios, such as passenger transportation and mining \cite{ref36}.

Parking path planning based on numerical optimization methods obtains the optimal path by minimizing or maximizing functions constrained by variables. 
The drawback of this method is that it is time-consuming and has low computational efficiency.
Therefore, it is not widely used in parking planning tasks and is usually employed for smoothing the initial trajectory or in scenarios emphasizing kinematic constraints.

In addition to path planning algorithms, the academic community has also explored using artificial intelligence algorithms to achieve automatic parking.
For example, a research group at Jiangsu University attempted to use BP neural networks to construct driver models to simulate parking behavior. 
Other researchers have proposed using genetic algorithms to train vehicles to park in a simulated environment \cite{ref39,lan2019simulated}. However, training neural networks is extremely time-consuming and requires a significant amount of computation.
Moreover, collecting a large amount of high-quality driving data necessary for training is challenging, and there are not enough widely recognized high-quality datasets available.
AI-based automatic parking controllers do not significantly outperform path planning-based automatic parking algorithms and are still constrained by the complex environment of parking garages, lacking robustness and generalizability \cite{ref28}. 
Therefore, in the development of a pure visual BEV perception planning scheme, it is unnecessary to prioritize AI parking methods.

Common traditional autonomous driving planning methods and their advantages and disadvantages are shown in \autoref{tab:plan_methods}. 
The proposed scheme in this paper is also included for comparison.

\section{Methodology}
\label{sec:methodology}

\subsection{Overall Architecture}
\label{sec:Overall_Workflow_of_the_Parking_Scheme}
\subsubsection{Overall Workflow}
\label{sec:Overall_Process}

The automatic parking planning method proposed in this paper can achieve parking path planning under the BEV local map.
Backing into a parking spot is a low-speed driving scenario, where the tires can be approximated as rigid bodies, and the influence of lateral forces need not be considered, allowing the application of vehicle kinematic models.
The main disadvantage of this model is that it allows instantaneous changes in steering angle, which is addressed in this paper using Bezier curves optimization to solve the problem of sudden changes in path curvature.

According to the analysis of various existing planning methods in \autoref{sec:Traditional_Planning_Algorithms}, it is found that: methods based on random sampling are ultimately unstable and easily affected by random factors.
Parking tasks in underground garages require high-precision trajectory planning, hence methods based on random sampling that cannot guarantee the rapid generation of optimal solutions are discarded.
Graph search algorithms, represented by A*, have the disadvantage of low efficiency and do not consider vehicle dynamics constraints, resulting in paths that may be discontinuous.
This shortcoming of discontinuous routes complements the advantages of interpolation curve planning.
The disadvantage of numerical optimization methods is their slow computation speed; however, for safety reasons, the last segment of the path when reversing into a garage needs to be driven slowly, and high planning precision is required.
Therefore, numerical optimization methods are well-suited for parking path planning, leveraging their strengths.

This paper proposes an automatic parking planning control method based on A* algorithm and numerical optimization methods, with the complete planning process as follows:

\begin{figure*}[!ht] \centering
    \includegraphics[width=0.7\linewidth]{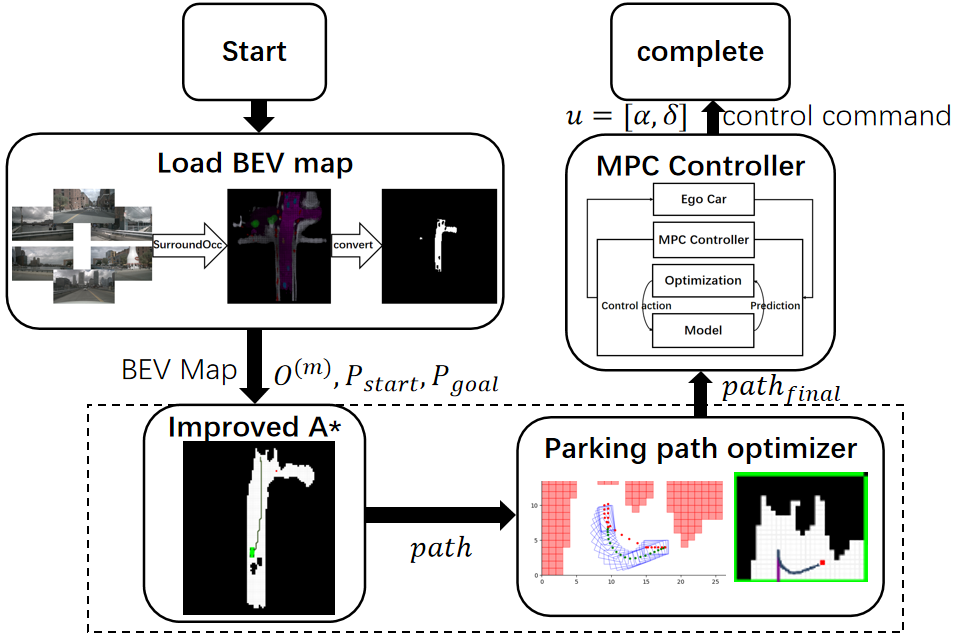}
    \caption{Overall architecture of the automatic parking planning method proposed in this paper.}
    \label{fig:pipeline}
\end{figure*}

\begin{enumerate}
    \item Load real environment information to occupy point clouds, and generate a BEV perspective grid map from this.
    \item Plan a path from the starting point to the endpoint based on the improved A* algorithm.
    The specific algorithm optimization ideas involved in this paper are detailed in \autoref{sec:Optimized_Path_Planning_Method_Based_on_A*_Algorithm}.
    \item Using the last segment of the path close to the parking spot as an initial value, generate a parking path using numerical optimization methods.
    The numerical optimization ideas used in this paper are detailed in \autoref{sec:Parking_Planning_Method_Based_on_Numerical_Optimization}.
    \item After the path is generated, perform simulation control.
    Collect path data and control command data as quantitative indicators to measure the quality of planning.
\end{enumerate}

See \autoref{fig:pipeline} for an illustration of the planning process.

\subsubsection{Kinematic Model}
\label{sec:Vehicle_Model}

To better simulate the motion of vehicles, a model of the vehicle is constructed.

\begin{enumerate}
    \item The vehicle does not exhibit vertical motion.
    \item Both front wheels have the same or approximately the same angle and rotational speed, and the same applies to both rear wheels.
    \item The front wheel steering angle controls the vehicle's yaw angle, while the rear wheels maintain the same direction as the vehicle body.
    \item The vehicle body and suspension are rigid bodies.
\end{enumerate}

\begin{figure}[!ht] \centering
    \includegraphics[width=0.95\linewidth,trim={150 140 150 145},clip]{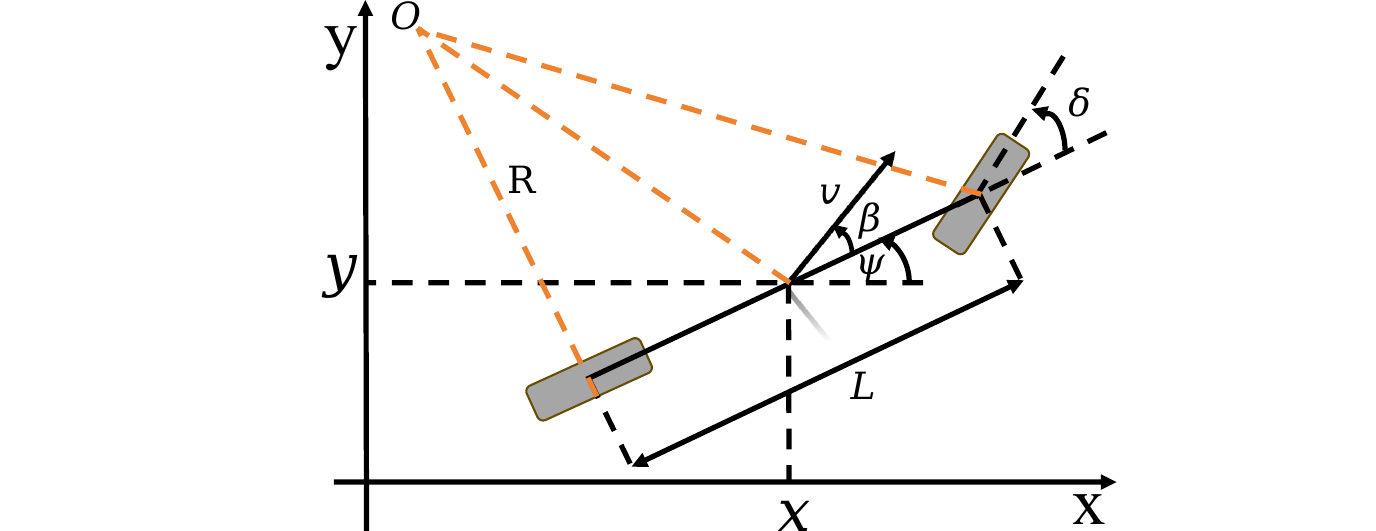}
    \caption{\label{fig:model}Vehicle Kinematic Model}
\end{figure}

Since each pair of front and rear wheels has the same state of motion, it is entirely possible to simplify the model by omitting each pair of front and rear wheels to just one wheel.
This leads to the classic bicycle model, which is suitable for describing the behavior of cars in low-speed scenarios \cite{ref17}.
The vehicle kinematic model used in this paper is shown in \autoref{fig:model}, which uses four quantities to describe the current state of the vehicle.

$x$: Horizontal ordinate of the vehicle's center

$y$: Vertical ordinate of the vehicle's center

$\psi$: Yaw angle of the vehicle

$v$: Speed of the vehicle

Additionally, the front wheel steering angle is denoted as $\delta$. 
According to the model assumptions, the car's steering angle is also this value.
Assuming the rear wheel steering angle is 0, meaning the rear wheels always face the same direction as the vehicle body. 
The slip angle refers to the angle between the direction of the vehicle's velocity and the orientation of the vehicle body, denoted as $\beta$.
Since parking is a low-speed scenario, $\beta$ can be considered extremely small and negligible, thus in this paper, $\beta=0$ is assumed. 
Furthermore, denote the vehicle's wheelbase as $L$, and the coordinates of the circle center corresponding to the steering trajectory during vehicle steering as $O$.

Based on the above model, the vehicle state vector can be obtained:
\begin{equation}
    z=\left[ x,y,v,\psi\right]
\end{equation}

As well as the vehicle state change model:
\begin{equation}
    \begin{cases} \dot{x}=  v \cdot  \cos (\psi)\\\dot{y} = v \cdot  \sin (\psi)\\\dot{v} = a\\\dot{\psi}= v \cdot  \tan (\delta)/L\end{cases} 
    \label{equation:car_model}
\end{equation}

According to \eqref{equation:car_model}, the control commands that the controller needs to output are given by \eqref{equation:control}:
\begin{equation}
    u=[a,\delta]
    \label{equation:control}
\end{equation}

At the same time, according to the Ackermann steering model, the minimum radius of the circular arc of the steering trajectory when the vehicle steers at its maximum steering capability can be calculated as:
$R=\frac{L}{\tan{\delta}}$.

\subsection{Optimized Path Planning}
\label{sec:Optimized_Path_Planning_Method_Based_on_A*_Algorithm}

This paper adopts an improved A* algorithm to complete navigation path planning.
The improved A* algorithm takes as input the BEV local grid map from upstream, reads the obstacle list, as well as the starting position $P_{start}=(x_{start},y_{start})$ and the parking spot location $P_{goal}=(x_{goal},y_{goal})$, to generate the entire navigation path from the starting point to the parking spot.

Compared to the traditional A* algorithm, the core improvement idea of this paper's method is to use heuristic optimization, binary heap optimization, and bidirectional search to reduce the number of searches and improve search efficiency. 
The introduction of the ego vehicle's volume enhances the feasibility of the generated path; then, Bezier curve optimization is used to smooth the path, ultimately improving the timeliness and trajectory quality of the A* algorithm, addressing the shortcomings of the A* algorithm such as discontinuous paths and large computational overhead and low efficiency.

\subsubsection{Ego Vehicle Volume}
\label{sec:Ego_Vehicle_Volume}

The A* algorithm starts with the starting point $P_{start}$ as the first current node $P_{current}$, and in each iteration, it expands the list of adjacent nodes for each current node $P_{current}$.
This involves determining the legality of adjacent nodes.
Each adjacent node $P_{adj}$ of the current node must be legal to be added to the open list $openlist$ and await further search.

There are two cases where an adjacent node $P_{adj}$ is illegal:

\begin{enumerate}
    \item $P_{adj}$ is already in the closed list $closelist$. This indicates that the node has been searched and does not need to be searched again.
    \item There is an obstacle at $P_{adj}$, and the ego vehicle cannot reach it.
\end{enumerate}

\begin{figure}[!ht] \centering
    \includegraphics[width=0.30\linewidth]{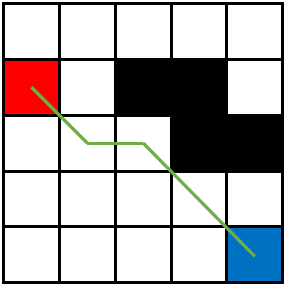}
    \caption{\label{fig:narrow}Trajectories that are too close to the wall or cross corners. These trajectories obviously cannot be used as parking trajectories in the real world.}
\end{figure}

Traditional A* algorithms abstract robots as points without volume.
However, in the real environment of automatic parking tasks, cars definitely have volume.
As shown in \autoref{fig:narrow}, paths generated by traditional A* algorithms may lead to dangerous situations, such as paths crossing corners or paths too close to walls.
These dangerous situations are unacceptable in automatic parking tasks.
If a car really drives along such a path, it will definitely result in scratches or collisions.
Therefore, this study needs to take into account the volume of the car.

Therefore, the improved A* algorithm extends the second illegal condition for $P_{adj}$ to: $P_{adj}$ and its vicinity have obstacles, and the ego vehicle cannot approach.

To determine whether a car can reach a node that does not have obstacles itself, it is necessary to further calculate the drivable area based on the obstacle map.
Because the car actually cannot reach areas that are too close to obstacles, when the center of the car reaches these areas, the outline of the car has already experienced scratches or collisions.

\begin{figure}[!ht] \centering
    \includegraphics[width=0.70\linewidth]{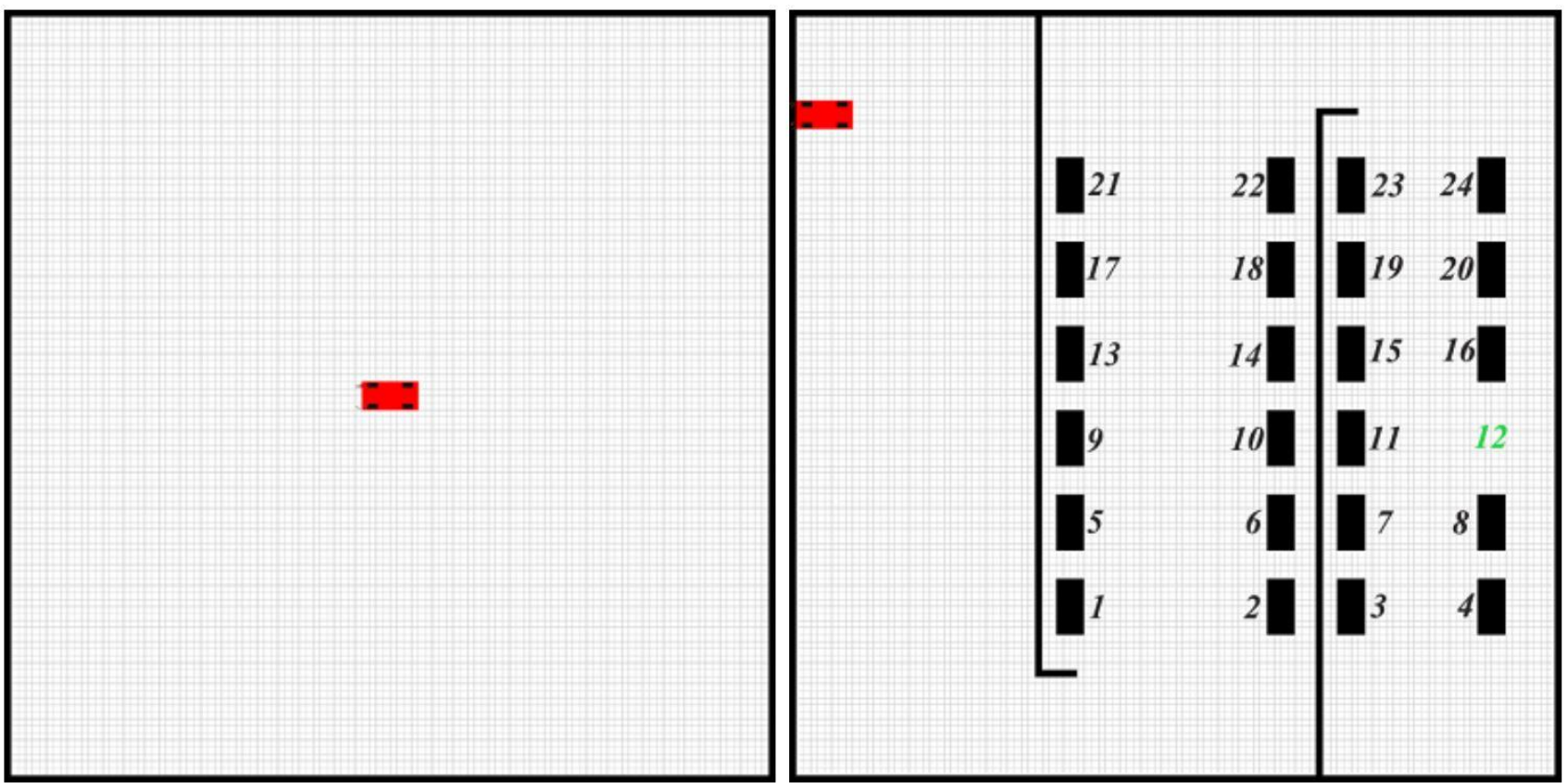}
    \caption{\label{fig:simple_map}Some papers and projects use similar maps for testing. Compared to real parking maps, the obstacles in these maps are too sparse.}
\end{figure}

The project that won first place in the Rahneshan Autonomous Driving Car Competition in 2020-2021 \cite{ref44} also took the volume of the ego vehicle into account in planning.
Their method is to traverse every location in the grid map when loading the map, compare it with the stored obstacle information, draw circles with the size of the car as the radius centered on each grid point, and determine whether there are obstacles near each location to calculate whether each location in the map allows the car to pass through.
The time complexity of this algorithm is $O(n^{3})$, which is theoretically very poor.
The reason this method can be used is that the project used obstacle-sparse maps as shown in \autoref{fig:simple_map}, and by reducing the scale of the data, the running speed was kept within a tolerable range.

\begin{figure} [!ht] \centering
    \includegraphics[width=0.5\linewidth,trim={50 80 70 50},clip]{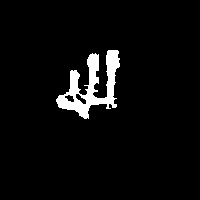}
    \caption{\label{fig:true_map}Grid map corresponding to a real scene}
\end{figure}

However, maps generated based on the real world contain far more obstacles than artificially made simulation maps, so the time consumption of the algorithm used in that project will rapidly inflate to an unacceptable level.
\autoref{fig:true_map} is a grid map corresponding to a real scene, which is a local map generated by the ego vehicle's perception module.
The black color signifies areas that are not drivable, and the white color signifies drivable roads.
It can be seen that in the automatic parking task scenario, the drivable area is very narrow, and most of the area is occupied by various obstacles.
Therefore, it is necessary to use more advanced algorithms to calculate the passability of nodes under the premise of introducing the volume of the ego vehicle.

This paper approximates the vehicle volume with a circle.
This can avoid vehicle collisions and make the generated path occupy more space than the width of the vehicle itself, leaving enough safety distance on both sides of the vehicle.
The necessary and sufficient condition for a node to be impassable is that there is an obstacle within the range of a circle centered on the node with a radius representing the size of the vehicle.

The proposed algorithm flow is roughly as follows:

\begin{enumerate}
    \item Initialize an obstacle matrix $O=(o_{ij}),i\in\left[ 0,height-1\right],y\in\left[ 0,width-1\right]$, where $height$ and $width$ are the height and width of the grid map, respectively.
    This matrix records whether each grid point of the grid map is allowed to pass through.
    Each element $o_{ij}$ of the matrix has three possible values: -1, 0, 1, indicating prohibited, unknown, and allowed, respectively.
    \item When expanding the adjacent node $P_{adj}=(x,y)$ of the current node, if $o_{xy}=1$, i.e., allowed to pass through, then add $P_{adj}$ to the $openlist$; if $o_{xy}=-1$, i.e., prohibited from passing through, then ignore this node.
    \item If $o_{xy}=0$, i.e., the current node is unknown whether it is allowed to pass through, then take $P_{adj}$ as the center, with the vehicle size $R_{size}$ as the radius, and traverse each point $P_{0}$ within the circle. Check these points for obstacles based on prior obstacle location information.
    If $P_{0}\in\mathbb{O}^{(m)}$, i.e., there is an obstacle at some point within the circle, then record this grid as prohibited from passing through on the obstacle matrix, assign $o_{xy}=-1$ and ignore this node; if there are no obstacles in the circular area, then record this grid as allowed to pass through on the obstacle matrix, assign $o_{xy}=1$, and add $P_{adj}$ to the $openlist$.
\end{enumerate}

At this point, it is possible to determine whether each adjacent node $P_{adj}$ of $P_{current}$ is legal, using $P_{current}$ and $\mathbb{O}^{(m)}$ as input information, to obtain the constantly updated obstacle matrix $O=(o_{ij})$ and $openlist$.

The advantage of this algorithm is obvious: by integrating the calculation of node passability with the A* algorithm search process, dynamic calculation of the obstacle matrix is successfully achieved.
This avoids calculating the passability of all nodes, greatly reduces the scale of the problem, and lowers the time complexity of the algorithm.

\subsubsection{Reduce Search Time}
\label{sec:Reduce_Search_Time}

By the method described in \autoref{sec:Ego_Vehicle_Volume}, the open list $openlist$ is obtained. The A* algorithm searches for points in the $openlist$ and adds them to the $closelist$ to gradually approach $P_{goal}$. This paper adopts heuristic optimization and bidirectional search to reduce the number of searches and improve search efficiency from a data structure perspective through binary heap optimization.

\paragraph{Heuristic Function Optimization}
\label{sec:Heuristic_Function_Optimization}

The core of the A* algorithm's ability to continuously expand nodes towards the goal and ultimately connect them into a complete path is the heuristic function that estimates the future cost of nodes.
The A* algorithm constantly selects the node with the lowest estimated cost from the $openlist$ as the current node $P_{current}$. 
Therefore, if the optimal node always has a significantly better cost, the A* algorithm can avoid repeatedly searching for multiple paths with similar costs and can keep moving forward along the optimal node without wasting time checking useless nodes.
This paper improves the speed of the algorithm in searching for the best node by optimizing the heuristic function. 
Many papers have tried to improve the performance of the A* algorithm by optimizing the heuristic function \cite{ref41,ref42,ref43}, but some works have not clearly explained their optimization principles, and the optimization effects are also limited.

In the physical world, cars have the ability to move in any direction, not just in the four directions of up, down, left, and right. 
Therefore, the Euclidean distance is chosen as the heuristic function to describe the estimated cost from the current node to the planned endpoint.
That is, for each node $P_{n}$ in the $openlist$, the heuristic cost function is defined as:
\begin{equation}
    h(n)=distance(P_{n},P_{goal})
\end{equation}
where$distance(P_{1},P_{2})=\sqrt{(x_{1}-x_{2})^{2}+(y_{1}-y_{2})^{2}} ,P_{i}=(x_{i},y_{i})$.
The actual cost function is:
\begin{equation}
    h(n)=distance(P_{current},P_{n})
\end{equation}

The A* algorithm uses the formula $f(n)=g(n)+h(n)$ to calculate the node cost. 
The node cost is essentially the sum of the actual cost function $g(n)$ and the heuristic function $h(n)$ in a 1:1 ratio. 
If the weight ratio of these two parts is adjusted, the behavior of the A* algorithm will change accordingly. 
Specifically, if the weight of the actual cost $g(n)$ is increased, the algorithm will be more inclined to consider the actual cost of the path already traveled, thus closer to the behavior of the Dijkstra algorithm.
When the weight of $g(n)$ is much greater than that of $h(n)$, the A* algorithm almost only considers the actual cost and does not consider the estimated cost, at this time, the A* algorithm has actually degenerated into the Dijkstra algorithm. 
Conversely, if the weight of the estimated function $h(n)$ is increased, the algorithm will be more inclined to rely on the estimated cost to guide the search direction.

Therefore, by adjusting the weight of the actual cost and the estimated cost, the search behavior of the A* algorithm can be effectively controlled to tend to consider either the actual cost or the estimated cost, thereby reducing unnecessary search points and improving search efficiency. 
Thus, the heuristic function is represented by the formula \eqref{equation:heuristic}:
\begin{equation}
    f(n)=g(n)+w \cdot h(n)
    \label{equation:heuristic}
\end{equation}

where $w$ is the weight coefficient, used to adjust the weight ratio of $g(n)$ and $h(n)$. 
In this way, the search tendency of the A* algorithm can be flexibly controlled by adjusting the value of $w$.

At the beginning of planning, the current node is relatively far from the endpoint, and there is no need for the algorithm to try all possible paths to obtain the global optimum. 
Instead, it is hoped that the algorithm can approach the endpoint as quickly as possible.
Therefore, during this stage, the algorithm uses a greedy strategy for search, completing the planning task with the lowest time consumption. 
At this time, increasing the weight is a good strategy to make the current node quickly converge to the endpoint.
However, when approaching the endpoint, it is hoped that the algorithm can find the best path to meet the requirements of fine planning for automatic parking tasks and improve the quality of the trajectory.

The specific implementation strategy is that when the Euclidean distance from the new node to the planned endpoint is less than a certain threshold, set $w<1$, hoping that the algorithm generates the optimal path; when the new node is far from the planned endpoint, set $w>1$, hoping that the path can quickly approach the endpoint. 
The specific weight values will be dynamically adjusted according to the experimental test results.

\begin{figure}[!ht]
    \centering
    \includegraphics[width=0.30\linewidth]{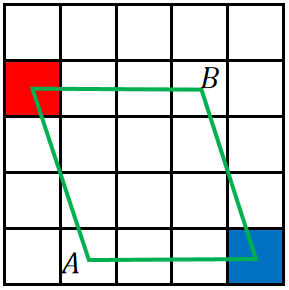}
    \caption{\label{fig:puzzled}Algorithm gets entangled when the cost of multiple paths is the same}
\end{figure}

A key factor leading to the performance degradation of the A* algorithm is that the algorithm gets entangled when encountering nodes with the same $f(n)$ value during node expansion.
When the cost $f(n)$ of multiple paths is the same, the algorithm searches these paths indiscriminately, even though in reality, the algorithm only needs to explore one of these paths.
As shown in \autoref{fig:puzzled}, the algorithm believes that the path from A and the path from B have the same cost, so it tries to plan from these two paths to the endpoint respectively. 
To improve this situation, the algorithm needs to make choices from nodes with the same heuristic function value to avoid unnecessary search overhead.

To achieve this choice, this project adds a small offset $p$ to the estimated cost $h(n)$. 
The role of the offset $p$ is to make priority judgments by comparing their estimated costs when the two nodes are the same.
Since nodes closer to the target endpoint usually have smaller estimated function values $h(n)$, adding a small offset $p$ to the estimated cost can ensure that nodes closer to the endpoint have slightly better heuristic function values $f(n)$, and thus are preferentially selected by the algorithm. 
At the same time, since the offset $p$ is very small, it will not significantly affect the established search tendency and interfere with the normal node search. 
Therefore, the node's cost function can be expressed in the form of formula \eqref{equation:final_heuristic}:
\begin{equation}
    f(n)=g(n)+(w+p) \cdot h(n)
    \label{equation:final_heuristic}
\end{equation}

In this way, the heuristic function can be further optimized, enabling the algorithm to avoid entanglement when faced with multiple nodes with the same cost, make decisive choices, reduce unnecessary search behavior, and further improve the search efficiency of the A* algorithm.

For each point $P_{n}\in openlist$, the algorithm can calculate its cost function value $f(n)$. Compared with the basic A*, the $f(n)$ calculated according to the method of this paper can better reflect the actual cost of the node and can flexibly adapt to the path planning requirements according to the distance from the endpoint. 
The above is the content of the method part of a computer science field autonomous driving thesis.

\paragraph{Bidirectional Search}
\label{sec:Bidirectional_Search}

The above content has improved the search efficiency of the A* algorithm, but the overall problem scale of the A* algorithm has not been reduced. 
Therefore, the A* algorithm is further optimized using a bidirectional search strategy.

Since the time consumption of the A* algorithm increases exponentially with the expansion of the problem scale, an optimization method that can effectively reduce the search space and improve algorithm performance is the bidirectional search algorithm \cite{lan2021learning,lan2021learning2,lan2022time}.
Bidirectional A* starts searching from both the start point and the goal point simultaneously, maintaining two current nodes $P_{current1}$ and $P_{current2}$, two open lists $openlist1$ and $openlist2$, and two closed lists $closelist1$ and $closelist2$.
For each direction of search, the scale of the problem that needs to be dealt with is greatly reduced. 
The general process of the algorithm is as follows:

\begin{enumerate}
    \item During initialization, $P_{current1}$ starts searching from $P_{start}$ towards $P_{goal}$, and $P_{current2}$ starts searching from $P_{goal}$ towards $P_{start}$.
    \item In each search process, each current node expands nodes with the lowest cost node in the other party's $closelist$ as the destination.
    \item During the search process, if $closelist1$ and $closelist2$ overlap, then from the overlapping point $P_{meet}$, recursively find the parent nodes towards $P_{start}$ and $P_{goal}$ to calculate the final path.
    \item If, during the search process, the $openlist$ of a certain $P_{current}$ is empty, meaning it cannot continue to expand, this implies that no path can be found between $P_{start}$ and $P_{goal}$, and the algorithm terminates.
\end{enumerate}

The bidirectional A* algorithm differs from the basic A* algorithm not only in the scale of the problem caused by unidirectional and bidirectional search but also in a significant difference: the target point of the bidirectional A* algorithm when expanding nodes changes constantly. 
The basic A* algorithm always calculates the estimated cost from $P_{n}$ to $P_{goal}$ when expanding nodes.
In contrast, the bidirectional A* algorithm uses the estimated distance from $P_{current}$ to the node with the lowest cost in the other direction as the heuristic cost when expanding nodes.
This difference causes $P_{current1}$ and $P_{current2}$ of the bidirectional A* algorithm to have a strong tendency to approach each other, which also ensures that when a path exists, the two segments of the path from $P_{start}$ to $P_{goal}$ and from $P_{goal}$ to $P_{start}$ of the bidirectional A* algorithm can definitely be connected into a complete path.

By searching from two directions, the improved A* algorithm maintains two $openlist$s, successfully reducing the problem scale and the number of node searches.

\paragraph{Binary Heap Optimization}
\label{sec:Binary Heap_Optimization}

After the above steps, the algorithm obtains $openlist$s from two directions and needs to search for $P_{best}$. 
This process requires the algorithm to traverse the entire $openlist$ multiple times to find $f_{min}$. 
Each traversal has a time complexity of $O(n)$, resulting in extremely long total time consumption. 
Therefore, if the time complexity of finding the optimal solution point in each operation can be reduced, theoretically, this project can improve the overall time efficiency of the algorithm by reducing the time consumption of finding the optimal node in the open list.

A binary heap, as a special tree structure, is stored in array form, with its top element always being the maximum or minimum value of the elements within the heap. 
The reason for choosing a binary heap is its ability to efficiently handle extreme value problems.

The insertion and deletion operations of a binary heap have a time complexity of $O(logn)$, and maintaining its relative order also only requires $O(logn)$ time.
Using a binary heap to store the open list can replace the operation of traversing the entire table with taking the top element of the heap and restructuring the heap, reducing the time complexity of finding the optimal node from $O(n)$ to $O(logn)$. 
This characteristic enables the binary heap to effectively reduce the time complexity of the A* algorithm.

Using binary heap optimization to reduce the time complexity of the A* algorithm's operation of selecting $P_{best}$ in each step.
The improved A* algorithm proposed in this paper uses a min-heap data structure to replace the list as the $openlist$ to store node data.
When selecting $P_{best}$, it is only necessary to simply take the top element of the heap and then restructure the min-heap, without the need to traverse the entire $openlist$. 
When adding a new node to the $openlist$, it is also not necessary to sort the entire $openlist$.

By using binary heap optimization, this paper can obtain $f_{min}$ from the $openlist$ of each direction with a better time complexity and obtain the corresponding $P_{best}$ as the new current node $P_{current}^{\prime}$.

\subsubsection{Bezier Curve Smoothing Optimization}
\label{sec:Bezier_Curve_Smoothing_Optimization}

This paper has discussed the optimization content applied in the planning process of the improved A* algorithm in \autoref{sec:Ego_Vehicle_Volume}-\autoref{sec:Binary Heap_Optimization}.
These optimizations reduce the time complexity of the algorithm while also taking into account the volume of the real vehicle. 
However, the algorithm is still based on grid map graph search, and the generated path still contains many sharp corners, with discontinuities at the junction of the planning path and the parking path. 
To solve these problems, after the path planning is completed, this paper applies Bezier curves to interpolate and optimize the path.

When $P_{current1}$ and $P_{current2}$ coincide, the algorithm starts from the coincidence point $P_{meet}$ and recursively finds the parent nodes towards $P_{start}$ and $P_{goal}$, and organizes them to obtain the original path $path_{origin}$.

The path $path_{origin}$ obtained directly using the A* algorithm on the grid map often exhibits characteristics of many line segments and large turning angles.
Clearly, cars in reality cannot complete sharp turns, so these line segments will inevitably have a significant impact on parking behavior. 
To enable cars to normally follow the trajectory, this method needs to smooth the $path_{origin}$. 
The purpose of curve smoothing is to eliminate unnecessary turns, reduce turning angles, optimize trajectory quality, ensure continuous curvature, and thus improve the usability of the path.

Bezier curves are a tool for fitting broken lines, which can generate a fitted curve entirely based on a series of control points.
Its general formula is:
\begin{equation} \small
    P(t)= \sum\limits_{i=0}^nP_iB_{i,n}(t),t\in\left[0,1\right]
\end{equation}

where $B_{i,n}(t)$ is the Bernstein basis function, expressed as follows:
\begin{equation} \small
    B_{i,n}(t)= \left(\begin{array}{c}n\\ i\end{array}\right) t^i(1-t)^{n-i},i\in\left[0,\cdots,n\right]
\end{equation}

While the Bezier method indeed has many advantages, it also has some shortcomings:

\begin{enumerate}
    \item The order of Bezier curves is directly limited by the number of vertices of the feature polygon. Once the number of vertices is too large, the computational load will increase rapidly.
    \item Bezier curves require high smoothness, which becomes relatively complex when splicing. Merging two Bezier curves is difficult to obtain a smooth new curve.
    \item Local modification cannot be achieved. Once a part of the curve is adjusted, the entire curve may be affected. This "chain reaction" characteristic is not ideal in some application scenarios.
\end{enumerate}

However, the B-spline method inherits all the advantages of the Bezier method while successfully overcoming the above three major shortcomings, making it more efficient and flexible in handling local modification requirements, complex curves, and large datasets.
B-spline curves can be considered an extended version of Bezier curves, or Bezier curves can be considered a special case of B-spline curves.

The general formula for B-spline curves is as follows:
\begin{equation} \small
    P(t)= \sum\limits_{i=0}^nP_iB_{i,k}(t)
\end{equation}
where
\begin{equation} \small
    \begin{array}{c}
    k=0, \quad B_{i, 0}(t)=\left\{\begin{array}{ll}
    1, & t \in\left[t_{i}, t_{i}+1\right] \\
    0, & \text { Otherwise }
    \end{array}\right. \\
    k>0, \quad B_{i, k}(t)=\frac{t-t_{i}}{t_{i+k}-t_{i}} B_{i, k-1}(t)\\+\frac{t_{i+k+1}-t}{t_{i+k+1}-t_{i+1}} B_{i+1, k-1}(t)
    \end{array}
\end{equation}

The formula for B-spline curves is very similar to that of Bezier curves, but it can be noted that the subscript has changed from the Bernstein basis function's $n$ to the B-spline basis function's $k$, indicating that the degree of the polynomial of the B-spline is unrelated to the number of control vertices, but is user-defined. 
At the same time, $t$ is no longer a continuous weight value, but a discrete list representing the position of control points. For example, the control point list $t$ can take values $\{0,\frac{1}{9},\frac{2}{9},\frac{3}{9},\frac{4}{9},\frac{5}{9},\frac{6}{9},\frac{7}{9},\frac{8}{9},1\}$, thus successfully dividing the curve into $m$ segments with $m+1$ nodes.

Using B-spline curves, this project can effectively smooth the path composed of multiple line segments generated by the A* algorithm, ensuring that the processed path is smooth and continuous.

According to the preset sampling rate, this paper selects $n$ points $P_{i}$ from the preliminary planned path $path_{origin}$ containing many unsmooth corners as the control points of the B-spline curve, where $i\in [0,n]$. 
Using these control points for B-spline fitting, a smooth path curve is obtained, and then the interpolated smooth curve is used as the final path $path$ generated by the planning algorithm.

By optimizing the $path_{origin}$ with Bezier curves, the final path $path$ can be obtained. 
The smoothness of the path generated by the improved A* algorithm has been significantly improved, enhancing the driving efficiency of cars in automatic parking tasks and the continuity and stability of trajectories.

\subsubsection{Inaccessible Parking Spots}
\label{sec:Inaccessible_Parking_Spots}

Throughout the text above, it has been assumed that the input data used for planning is legal and correct. 
However, this may not always be the case in reality, and this research needs to reserve fault tolerance for illegal input data.
Due to the narrow of underground garage environments, the automatic parking planning module often encounters situations where the parking spot is unreachable.
Incorrect berth coordinates output by the upstream module, or too many obstacles near the berth that do not leave enough space for the car to complete the reverse parking operation, may result in the car being unable to park in the berth.
It is essential to design a strategy that allows the planning module to attempt to make a parking plan even when the parking spot is unreachable.

This paper implements a method to attempt planning when the parking spot is unreachable by maintaining a temporary optimal node during the A* algorithm search process. 
While expanding new nodes in the forward search, the algorithm will determine whether this new node is closer to the planning endpoint than the previous nodes. 
If the expanded new node is closer to the planning endpoint, it will be updated as the temporary optimal node. 
If the planning algorithm fails to search for a reachable path, the algorithm essentially degenerates into a unidirectional A* search, continuously moving forward until it finds an optimal temporary node.
The algorithm will then use the temporary optimal node as the endpoint, recursively tracing back its parent nodes until it returns to the starting node.
In this way, the algorithm can output a path that is as close as possible to the planning endpoint and navigate the car to a location close to the endpoint.

The implementation of this feature allows the planner to output a path even in the case of illegal parking spots, rather than waiting indefinitely, enhancing the algorithm's ability to cope with special situations.

\subsubsection{Summary}
\label{sec:method_Summary}

This paper discusses the specific optimization content of the improved A* algorithm in detail from \autoref{sec:Ego_Vehicle_Volume} to \autoref{sec:Inaccessible_Parking_Spots}, with the following specific improvements:

\begin{enumerate}
    \item Introducing the ego vehicle's volume to make the generated path more realistic.
    \item Reducing search overhead and improving algorithm computing speed and real-time performance through heuristic function optimization, bidirectional search, and binary heap optimization.
    \item Applying Bezier curve optimization to achieve path smoothing.
    \item Considering the unreachable parking spot situation to increase algorithm robustness.
\end{enumerate}

Combining the above optimization methods, the A* algorithm proposed in this paper can generate higher-quality continuous paths faster, achieving navigation path planning from the starting point to the parking spot.

\subsection{Parking Planning Optimizer}
\label{sec:Parking_Planning_Method_Based_on_Numerical_Optimization}

Improved A* algorithm can quickly generate a navigation path from the starting point to the endpoint. However, it cannot achieve fine planning for reverse parking.
This is a drawback of graph search algorithms—planning accuracy is limited by the grid size. 
Furthermore, infinitely increasing grid density would lead to an unacceptable trade-off in planning time. 
Therefore, a numerical optimization-based method is introduced to complete the last section of the reverse parking path planning.

This paper adopts a numerical optimization method to output the final parking path $path_{final}$ based on the navigation path $path$ generated by the improved A* algorithm. 
Optimizer reads the information of the last section of the navigation path, generates a simple geometric curve as the initial value, and then uses numerical optimization methods to optimize the initial value into a feasible parking path.
The optimized path ensures compliance with vehicle kinematic constraints and obstacle constraints.

\subsubsection{Optimization problem modeling}
\label{sec:Problem_Formulation}

The work that the optimizer needs to complete can be simply described as follows: Given the initial state and the final state (the state of completing parking), considering all obstacles and vehicle dynamics constraints, find a series of control commands $u$ that allow the vehicle to move from the initial state to the final state, and make the driving trajectory as optimal as possible. 
Next, construct the constraint model based on the known information.

The representation of the vehicle state refers to \autoref{sec:Vehicle_Model}.
Vehicle state is expressed as a vector $z=\left[ x,y,v,\psi\right]$ containing four dimensions. 
When introducing the optimizer into the parking planning system, these four dimensions are still used to express the vehicle's state, namely: the lateral coordinate of the center of the rear axle, the longitudinal coordinate of the center of the rear axle, the vehicle speed, and the vehicle yaw angle.
In the road section processed by the optimizer, the algorithm can obtain the initial state information from the output result of the improved A* algorithm, and can obtain the position and pose of the final state according to the parking spot information, while the speed of the final state is naturally 0.

To conform to the writing habits of optimization problems, in the subsequent content of \autoref{sec:Problem_Formulation}, $x$ will be used to represent the vehicle's state.
In addition, $u$ is used to represent the control command, and $f$ represents the dynamics of the system. 
Then, the state transition can be described by the following formula at time $k$:
\begin{equation} \small
    x_{k+1}=f(x_{k},u_{k})
    \label{equation:state_trans}
\end{equation}

The vehicle state and control commands themselves are also subject to constraints.
First, the vehicle cannot cross the map boundary, nor can it race at unusually fast speeds. 
It is obviously impossible for the vehicle to have rocket-like acceleration or to turn around in place in an instant. 
The following expression can be used to represent the constraints that the vehicle state and control commands must satisfy at each moment.
\begin{equation} \small
    h(x_{k},u_{k})\leq0
    \label{equation:bound_cons}
\end{equation}

Considering that the car is a three-dimensional object, it is treated as a rectangle. 
At state $x_{k}$, $\mathbb{E}(X_{k})$ is used to represent the space occupied by the car in its current state. 
$\mathbb{E}(X_{k})$ can be represented as follows. 
Where $R\left(x_{k}\right)$ and $t(x_{k})$ define the rotation and translation of the vehicle in space in matrix form, respectively.
\begin{equation} \small
\mathbb{E}\left(x_{k}\right)=R\left(x_{k}\right) \mathbb{B}+t\left(x_{k}\right), \quad \mathbb{B}:=\left\{y: G y \preceq_{\bar{\kappa}} g\right\}
\end{equation}

This method hopes that at any state of the planned trajectory, the vehicle will not collide with any obstacle.
Let $\mathbb{O}^{(m)}$ represent the $m$-th obstacle in the obstacle set containing $M$ obstacles.
Then, the collision constraint can be expressed as follows:
\begin{equation} \small
    \mathbb{E}\left(x_{k}\right) \cap \mathbb{O}^{(m)}=\emptyset, \quad \forall m=1, \ldots, M
    \label{equation:collision_cons}
\end{equation}

Treating obstacles in space as convex sets, we have:
\begin{equation} \small
    \mathbb{O}^{(m)}=\left\{y \in \mathbb{R}^{n}: A^{(m)} y \preceq \mathcal{K} b^{(m)}\right\}
\end{equation}

The parking path generator based on numerical optimization methods proposed in this paper incorporates \eqref{equation:state_trans}\eqref{equation:bound_cons}\eqref{equation:collision_cons} into the constraint function.

The objective function used by the parking trajectory optimizer is divided into four parts:

\begin{enumerate}
    \item The vehicle state should be as close as possible to the desired state.
    \item The desired control quantity should be as small as possible.
    \item The rate of change of the control quantity should be as small as possible.
    \item The first control quantity should be as close as possible to the initial value's acceleration and steering angle. The initial value's acceleration and steering angle are represented by $u_{start}$.
\end{enumerate}

Combining the above four parts, $w_1$, $w_2$, $w_3$, and $w_4$ are selected as the weights for the four items, resulting in the objective function of the parking trajectory optimization problem as formula \eqref{equation:object_function}:

\noindent
\begin{equation} \small
\begin{split}
    l(x(k), u(k))=w_{1}\left\|x(k)-x_{r e f}(k)\right\|_{2}^{2}+w_{2}\|u(k)\|_{2}^{2}+ \\
w_{3}\left\|\frac{(u(k)-u(k-1))}{t_{s}}\right\|_{2}^{2}+w_{4}\|u(0)-u_{start}\|_{2}^{2}
\end{split}
    \label{equation:object_function}
\end{equation}

Using this objective function, the optimizer can generate a parking trajectory from the starting state to the parked position in the parking space with as high precision and as low cost as possible.

\subsubsection{Parking Trajectory Generation}
\label{sec:Parking_Path_Generation}

This paper uses the geometric curve method to generate a rough parking path as the initial Guess for the optimizer. 
This rough path does not fully consider the vehicle body volume and obstacle avoidance issues, nor does it consider kinematic constraints beyond the vehicle's minimum turning radius. 
This method selects a short section of the path close to the endpoint from the planning results of the improved A* algorithm, and then calculates the geometric curve based on the start and end points of this section of the path. 
By using the geometric curve method, a nearly approximate path is quickly generated and used as the input for the numerical optimizer. 
The optimizer then corrects it into a smooth parking path that meets kinematic constraints and obstacle constraints.

The optimizer establishes an optimization problem model based on the constraints in \autoref{sec:Problem_Formulation}, considering the vehicle dynamics model, obstacle constraints, and the vehicle's own volume.
After the model is established and constraints and objective functions are imported, IPOPT (Interior Point OPTimizer) can be called to solve this nonlinear optimization problem. 
IPOPT can solve optimization problems in the form of \eqref{equation:Format_requirement}. Where $f(x): R^{n} \rightarrow R$ represents the objective function of the optimization problem. 
$g(x): R^{n} \rightarrow R^{m}$ represents the constraint function. 
$g_{L}$ and $g_{U}$ represent the boundaries of the constraint function $g(x)$. 
$x_{L}$ and $x_{U}$ represent the boundaries of the optimization variable $x$.
\begin{equation} \small
    \begin{array}{l}
    \underbrace{\min }_{x \in R^{n}} f(x) \\
    
    \begin{split}
    &s.t.\quad  \left\{\begin{array}{lc}
     g_{L} \leq g(x) \leq g_{U} \\
    x_{L} \leq x \leq x_{U}
    \end{array}\right.
    \end{split}
    
    \end{array}
    \label{equation:Format_requirement}
\end{equation}

By dualizing the obstacle constraints, the parking planning problem can be transformed into a similar format \cite{ref45}.
The final optimization problem is represented by the formula \eqref{equation:final_form}:
\begin{equation} \small
\begin{array}{cl}
\mathop{min}\limits_{\mathbf{x}, \mathbf{u}, \mathbf{s}, \boldsymbol{\lambda}, \boldsymbol{\mu}} & \sum_{k=0}^{N}\left[\ell\left(x_{k}, u_{k}\right)+\kappa \cdot \sum_{m=1}^{M} s_{k}^{(m)}\right] \\
\text { s.t. } & x_{0}=x_{S}, x_{N+1}=x_{F}, \\
& x_{k+1}=f\left(x_{k}, u_{k}\right), h\left(x_{k}, u_{k}\right) \leq 0, \\
& -g^{\top} \mu_{k}^{(m)}+\left(A^{(m)} t\left(x_{k}\right)-b^{(m)}\right)^{\top} \lambda_{k}^{(m)}>-s_{k}^{(m)}, \\
& G^{\top} \mu_{k}^{(m)}+R\left(x_{k}\right)^{\top} A^{(m)^{\top}} \lambda_{k}^{(m)}=0, \\
& \left\|A^{(m)^{\top}} \lambda_{k}^{(m)}\right\|_{*}=1, \\
& s_{k}^{(m)} \geq 0, \lambda_{k}^{(m)} \succeq \mathcal{K}^{*} 0, \mu_{k}^{(m)} \succeq \overline{\mathcal{K}}^{+} 0 \\
& \text { for } k=0, \ldots, N, m=1, \ldots, M .
\end{array}
    \label{equation:final_form}
\end{equation}

\begin{figure}[!ht]  \centering
    \includegraphics[width=0.75\linewidth]{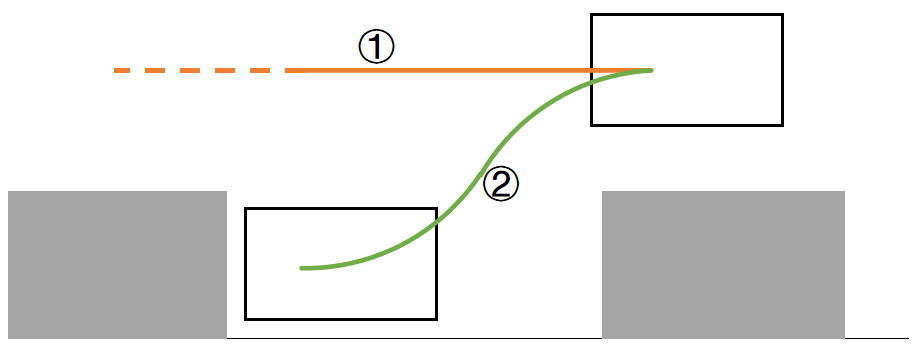}
    \caption{\label{fig:parking_pipeline}Parking process described in this paper. The orange path \ding{172} represents the navigation path planned by the improved A* algorithm. The green path \ding{173} represents the reverse parking path generated by the numerical optimization method using the data of the last section of the path planned by the improved A* algorithm combined with the geometric curve as input. The two paths are seamlessly connected to form a complete path.}
\end{figure}

The output of the optimizer is the planned parking path information $path_{final}$ for reverse parking, i.e., the $x$ in formula \eqref{equation:Format_requirement}, which is used for the car to finally complete the parking into the parking space. 
It includes acceleration (representing engine output) and steering angle (representing steering wheel angle), as well as the position of each point on the path.
The numerical optimization method ensures that this path is collision-free and complies with vehicle kinematic constraints.
As shown in \autoref{fig:parking_pipeline}, the final generated path is seamlessly spliced from the navigation path planned by the improved A* algorithm and the parking path planned by the numerical optimizer. 
The car drives along the navigation path generated by the improved A* algorithm to the vicinity of the parking spot, and then follows the parking path to complete the reverse parking action.

\subsection{Main Contributions}
\label{sec:Main_Contributions_of_This_Paper}

This paper proposes a targeted planning scheme for the task of local map autonomous parking, aiming to meet the requirements of real-time performance and safety for this task, while also improving the comfort of the planned path.

\begin{enumerate}
    \item Optimize the A* algorithm to improve its planning speed and resolve issues of path discontinuity and excessive turns, in order to meet the high real-time performance and high trajectory quality requirements of BEV local map path planning.
    \item Use optimizer to find obstacle-free parking trajectories in local maps, ensuring good obstacle avoidance safety.
    \item Carry out several sets of comparative experiments in simulator, prove that the proposed planning method is superior to the traditional A* algorithm combined with geometric curve methods in terms of real-time performance, safety, and comfort.
\end{enumerate}

\section{Experiments}
\label{sec:experiments}

\subsection{Experimental setup}
\label{sec:Test_Environment}

The environment used for testing the autonomous parking planning and control scheme in this paper consists of two parts: the hardware platform and the simulation system. 
The hardware platform includes a personal PC workstation, and the simulation system includes a Python-based test environment and the CARLA-ROS simulation environment.

\subsubsection{Hardware Platform}
\label{sec:Hardware_Platform}

The simulation experiments are implemented on the Ubuntu 20.04 LTS operating system based on the Linux kernel. 
The computer is equipped with one CPU, 8 cores, and 16 threads, with the CPU model being Intel(R) Core(TM) i7-10700KF CPU @ 3.80GHz.
The PC has a total of 32GB of available memory space. 
The graphics card used is a GeForce RTX 3070.

\subsubsection{CARLA-ROS Joint Simulator}
\label{sec:CARLA-ROS_Joint_Simulation_Environment}

\begin{figure*}[!ht]  \centering
    \includegraphics[width=0.95\linewidth]{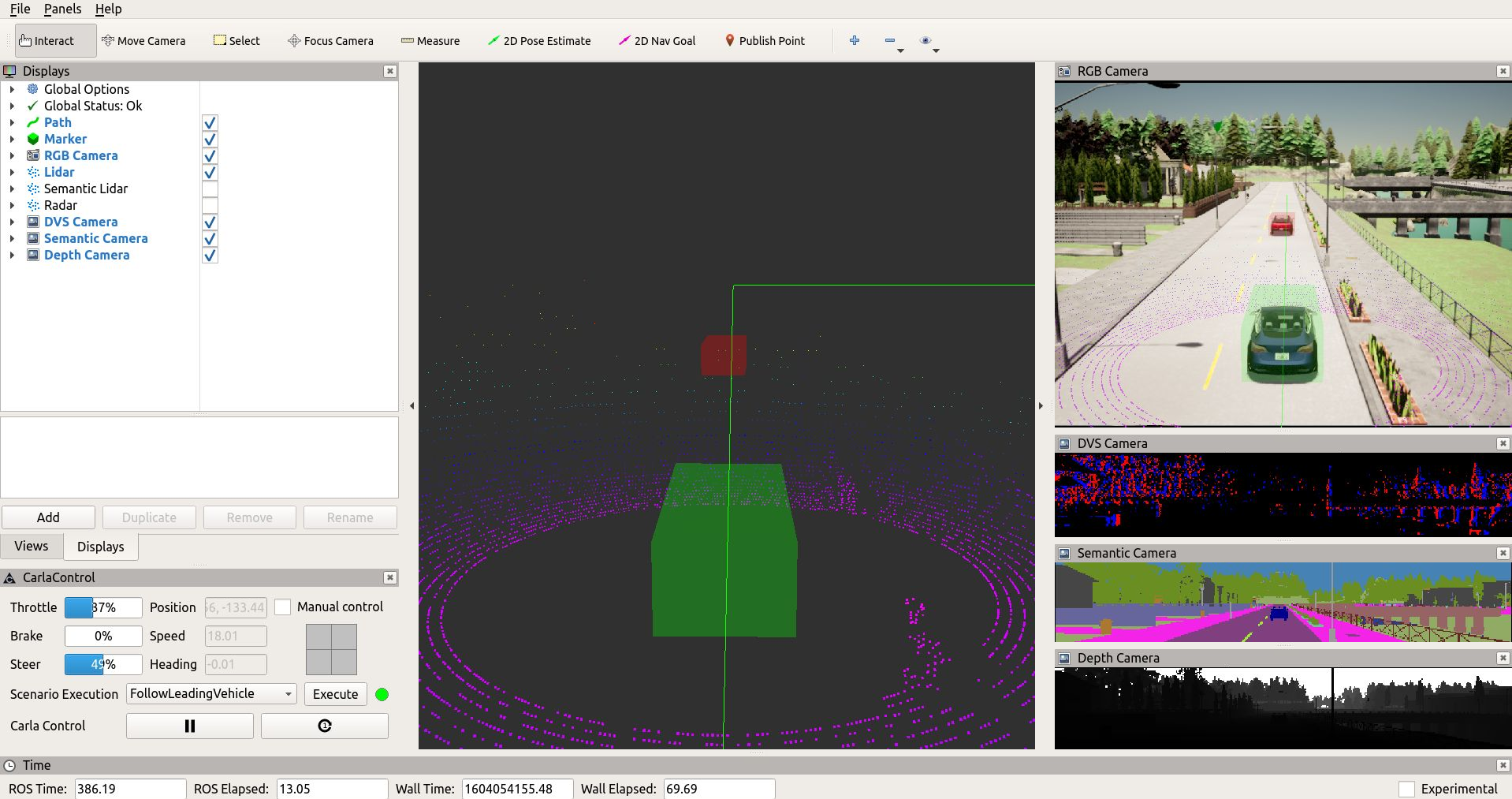}
    \caption{Our customized CARLA-ROS Simulator.}
    \label{fig:carla}
\end{figure*}

The CARLA simulator is widely recognized for its high customizability and physical realism, making it a common choice for autonomous driving simulations. 
This study employs a joint simulation environment integrating CARLA with the widely-used robotic development environment ROS as the simulation test environment.
Bridging the CARLA simulator with ROS facilitates the development, testing, and validation of autonomous vehicles on a unified platform.

\subsubsection{Python Environment}
\label{sec:Python_Test_Environment}

The algorithm implemented in this paper is written in Python 3.8.18, using VS Code as the IDE, and the code working environment is managed with conda 23.10.0. 
The framework used is Pytorch 1.10.1+cu111. The simulation system is built using Python, employing the CARLA-ROS joint simulation environment for simulation testing, with test data and commands interacting with the Python simulation environment through the Ackermann Control Command package.

\begin{figure*}[!ht]  \centering
    \includegraphics[width=0.95\linewidth]{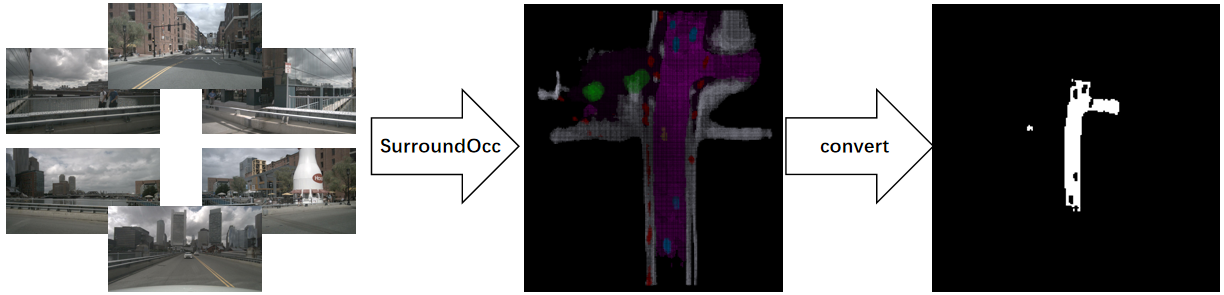}
    \caption{\label{fig:map_pipeline}Grid map acquisition process. Multi-camera images are converted to point clouds by SurroundOcc, then the point clouds are cleaned and converted to grid maps.}
\end{figure*}

This paper uses maps generated from real scenarios as the grid maps for the autonomous parking planning algorithm. 
The map size is 200$\times$200.SurroundOcc \cite{ref46} perception model developed by Tsinghua University and Tianjin University generates BEV occupancy point clouds from multi-camera images are provided by the nuScenes dataset. 
Then remove outliers and occlusion points, convert point clouds into grid maps for simulation testing.
A schematic diagram of converting occupancy point clouds to grid maps is shown in \autoref{fig:map_pipeline}.

\subsection{Algorithm Time Testing}
\label{sec:Algorithm_Time_Testing}

To compare the performance of the parking planning algorithm based on the improved A* algorithm and numerical optimization proposed in this paper, multiple sets of simulation comparative tests are conducted.
All experiments are under the same conditions, to ensure the fairness and comparability of the results.

Test navigation algorithm planning speed in the simulator to verify its real-time performance. 
Record the time when the improved A* algorithm starts execution, and timing ends when the improved A* algorithm completes planning and finds a complete path from the starting point to the parking spot. 
Since this study mainly focuses on the real-time performance of the navigation path planning algorithm, subsequent steps such as generating geometric curves and calculating the final parking path using the optimizer are not included.

The complete time frame can be divided into five segments: initialization, obstacle map establishment, heuristic search for paths, backtracking path generation, and interpolation optimization of paths.
Initialization time consists of time consumption of reading grid map images, creating parking simulator, completing vehicle model and controller initialization.
Obstacle map establishment involves generating reachable areas for vehicles based on obstacle information and vehicle volume.
Path planning time, involves the time consumption for the improved A* algorithm to find a path from the starting point to the parking spot.
Backtracking path generation and interpolation optimization of paths have a very small proportion of total time consumption, which this paper does not focus on.

This study conduct the comfort verification mentioned in \autoref{sec:Comfort_Testing} by examining the average acceleration and average steering wheel angle of the driving trajectory.

The following metrics are used to evaluate the performance of the planning algorithm:
\begin{enumerate}
    \item Initialization time $(ms)$
    \item Obstacle map loading time $(ms)$
    \item Planning path time $(ms)$
    \item Average trajectory acceleration $(m/s^{2})$
    \item Average steering wheel turning angle $(\degree)$
\end{enumerate}

\subsection{Dangerous Scenario Testing}
\label{sec:Dangerous_Scenario_Testing}

A series of parking scenarios are designed, to test the proposed parking path optimizer.
These scenarios vary in their level of danger and can be effectively used to evaluate the planning effectiveness of the parking path planner under different conditions.

For each scenario, this paper will present the grid map near the parking spot and the corresponding real-world image. 
In \autoref{fig:caseA} to \autoref{fig:caseF}, (a) shows the grid map, and (b) shows the real-world image. 
In the grid map, the blue border represents the preset parking position, and the pink area represents the obstructed area output by the upstream perception module that cannot be entered. Each grid square represents a meter.

\subsubsection{Scenario A}

\begin{figure}[ht!]
  \centering
    \begin{subfigure}[t]{0.18\textwidth}
      \centering   
      \includegraphics[height=2.9cm,trim={0 0 25 28},clip]{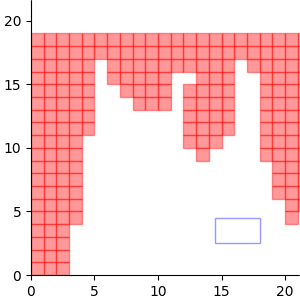}
        \caption{Grid map near the parking spot}
        \label{fig:sub.A1}
    \end{subfigure}
    \begin{subfigure}[t]{0.3\textwidth}
      \centering   
      \includegraphics[height=2.9cm]{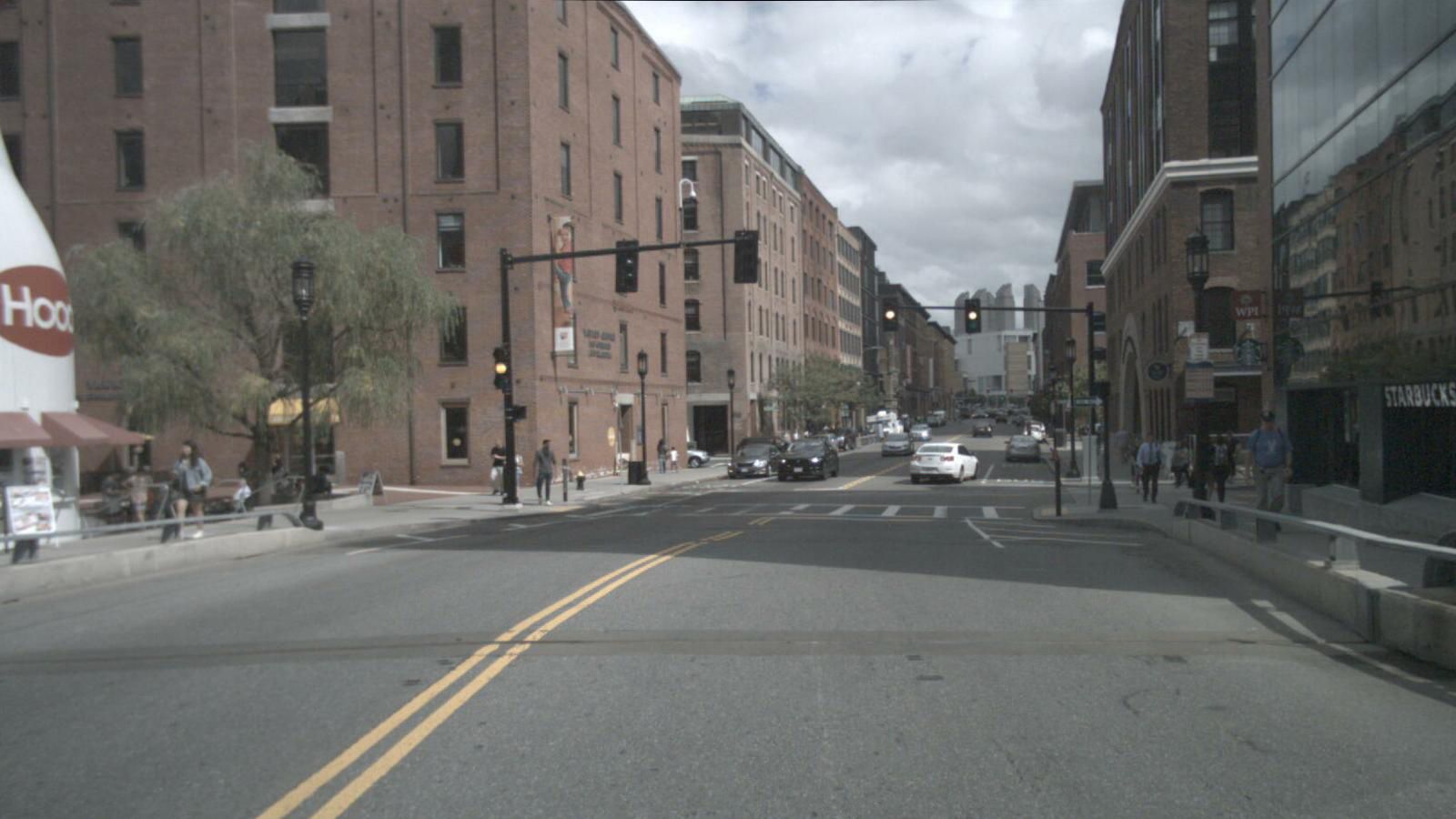}
        \caption{Real scene}
        \label{fig:sub.A2}
    \end{subfigure}
\caption{\label{fig:caseA}Scenario A, traffic flow intersection}
\end{figure}

As shown in \autoref{fig:caseA}, requires the vehicle to depart from the bottom of the grid map, reach the parking spot, and park vertically.
Upon reviewing the real-world , we find that the parking spot is located on a spacious roadside, but attention needs to be paid to the nearby traffic flow.

\subsubsection{Scenario B}
\begin{figure}[ht!]
  \centering
    \begin{subfigure}[t]{0.175\textwidth}
      \centering   
      \includegraphics[height=2.9cm]{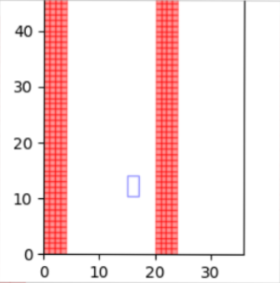}
        \caption{Grid map near the parking spot}
        \label{fig:sub.B1}
    \end{subfigure}
    \begin{subfigure}[t]{0.3\textwidth}
      \centering   
      \includegraphics[height=2.9cm]{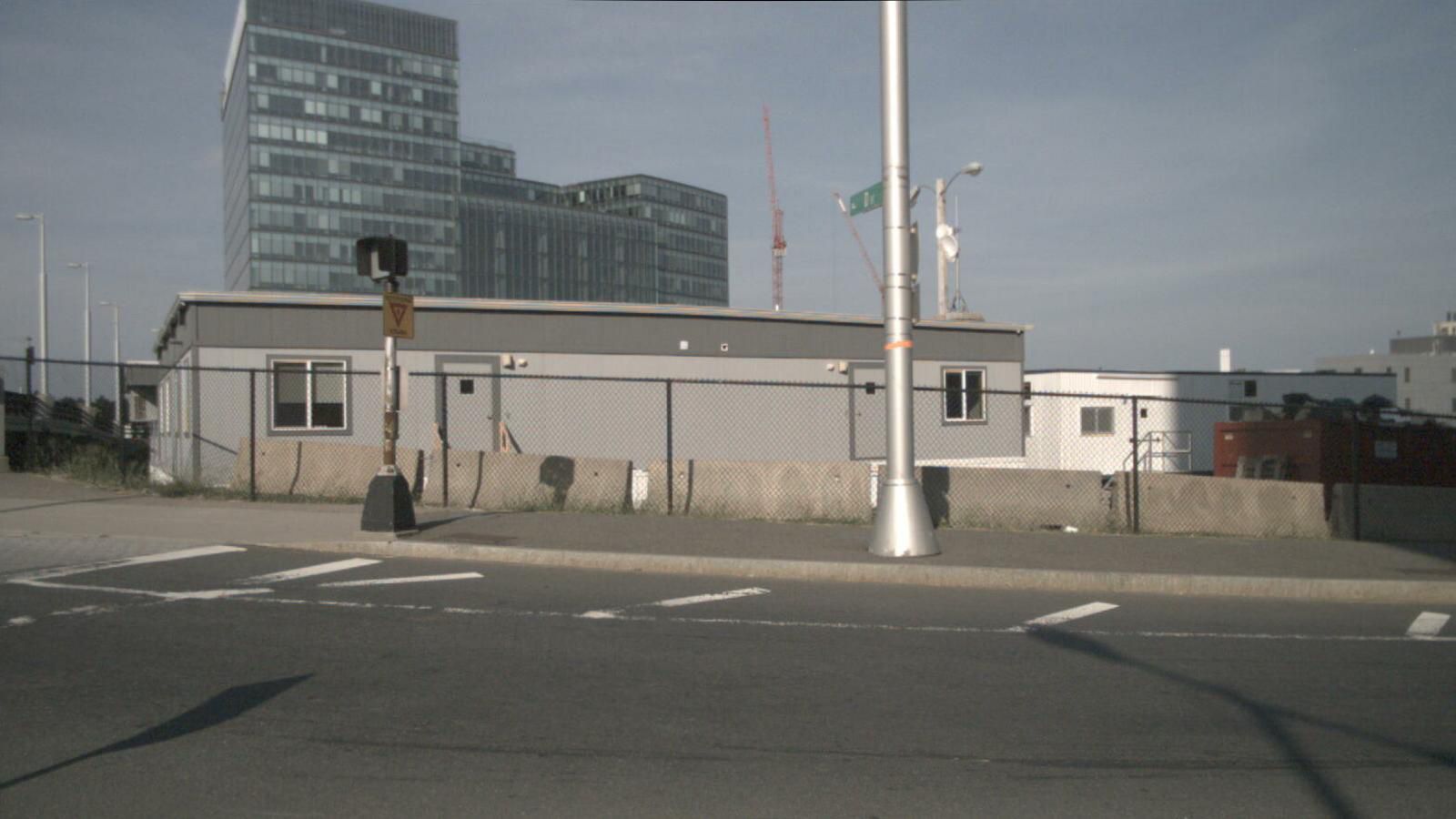}
        \caption{Real scene}
        \label{fig:sub.B2}
    \end{subfigure}
\caption{\label{fig:caseB}Scenario B, ideal situation}
\end{figure}

Depicted in \autoref{fig:caseB}, involves the vehicle approaching the parking spot from the bottom of the grid map and then parking parallel to the road on the right side. 
This scenario is spacious with no obstacles near the parking spot or on the traffic route, making it an ideal parallel parking situation.

\subsubsection{Scenario C}

\begin{figure}[ht!]
  \centering
    \begin{subfigure}[t]{0.18\textwidth}
      \centering   
      \includegraphics[height=2.9cm,trim={0 0 20 20},clip]{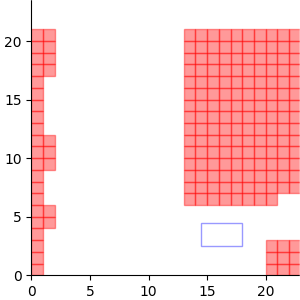}
        \caption{Grid map near the parking spot}
        \label{fig:sub.C1}
    \end{subfigure}
    \begin{subfigure}[t]{0.3\textwidth}
      \centering   
      \includegraphics[height=2.9cm]{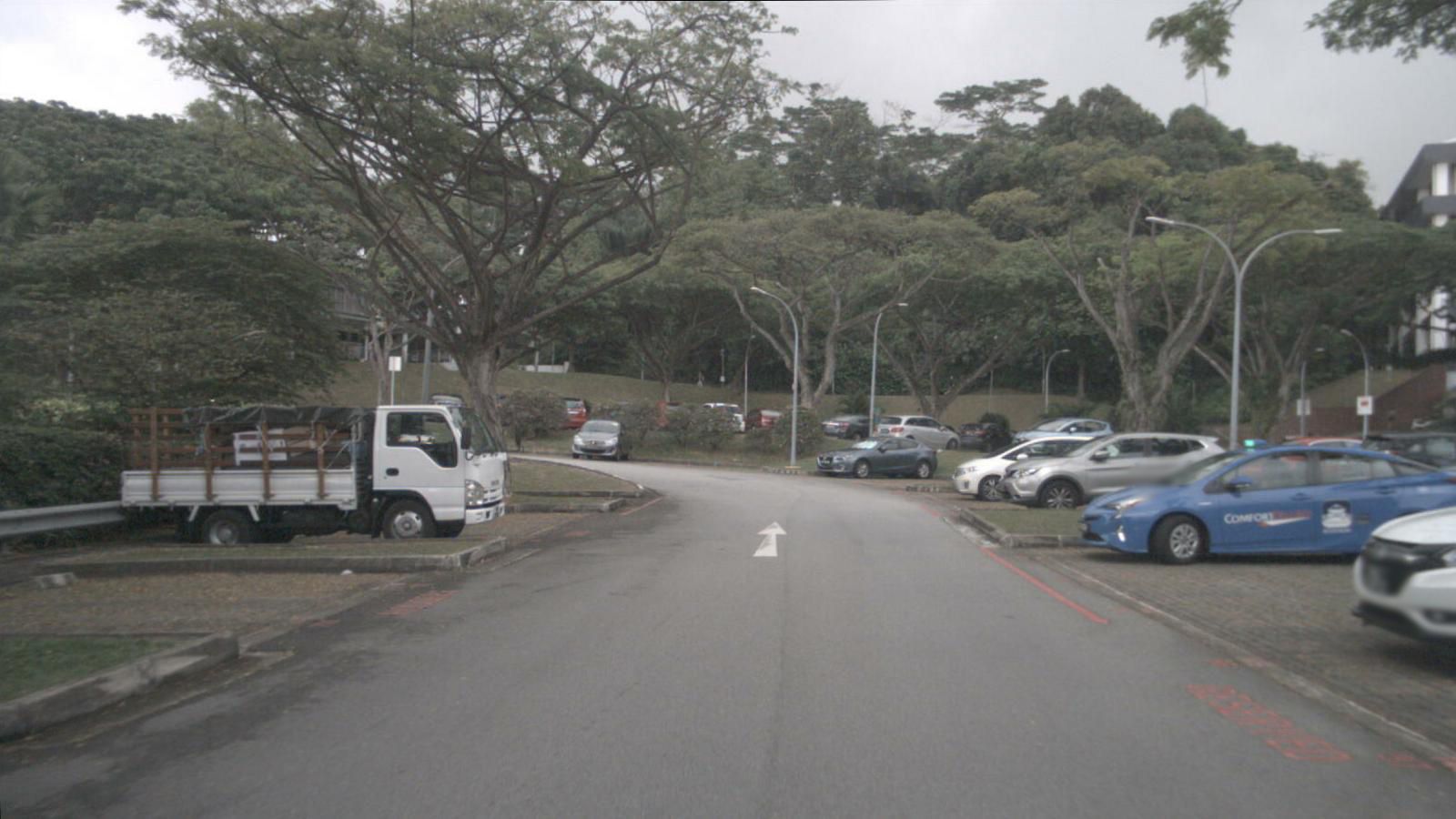}
        \caption{Real scene}
        \label{fig:sub.C2}
    \end{subfigure}
\caption{\label{fig:caseC}Scenario C, close to one side}
\end{figure}

Illustrated in \autoref{fig:caseC}, requires the vehicle to approach from below and park in a parking spot on the right side of the parking lot road.
While the parking spot is spacious, one side is occupied by other parked vehicles.

\subsubsection{Scenario D}
\begin{figure}[ht!]
  \centering
    \begin{subfigure}[t]{0.17\textwidth}
      \centering   
      \includegraphics[height=2.9cm,trim={0 0 25 25},clip]{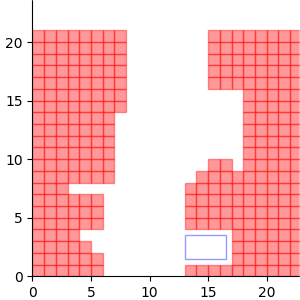}
        \caption{Grid map near the parking spot}
        \label{fig:sub.D1}
    \end{subfigure}
    \begin{subfigure}[t]{0.3\textwidth}
      \centering   
      \includegraphics[height=2.9cm]{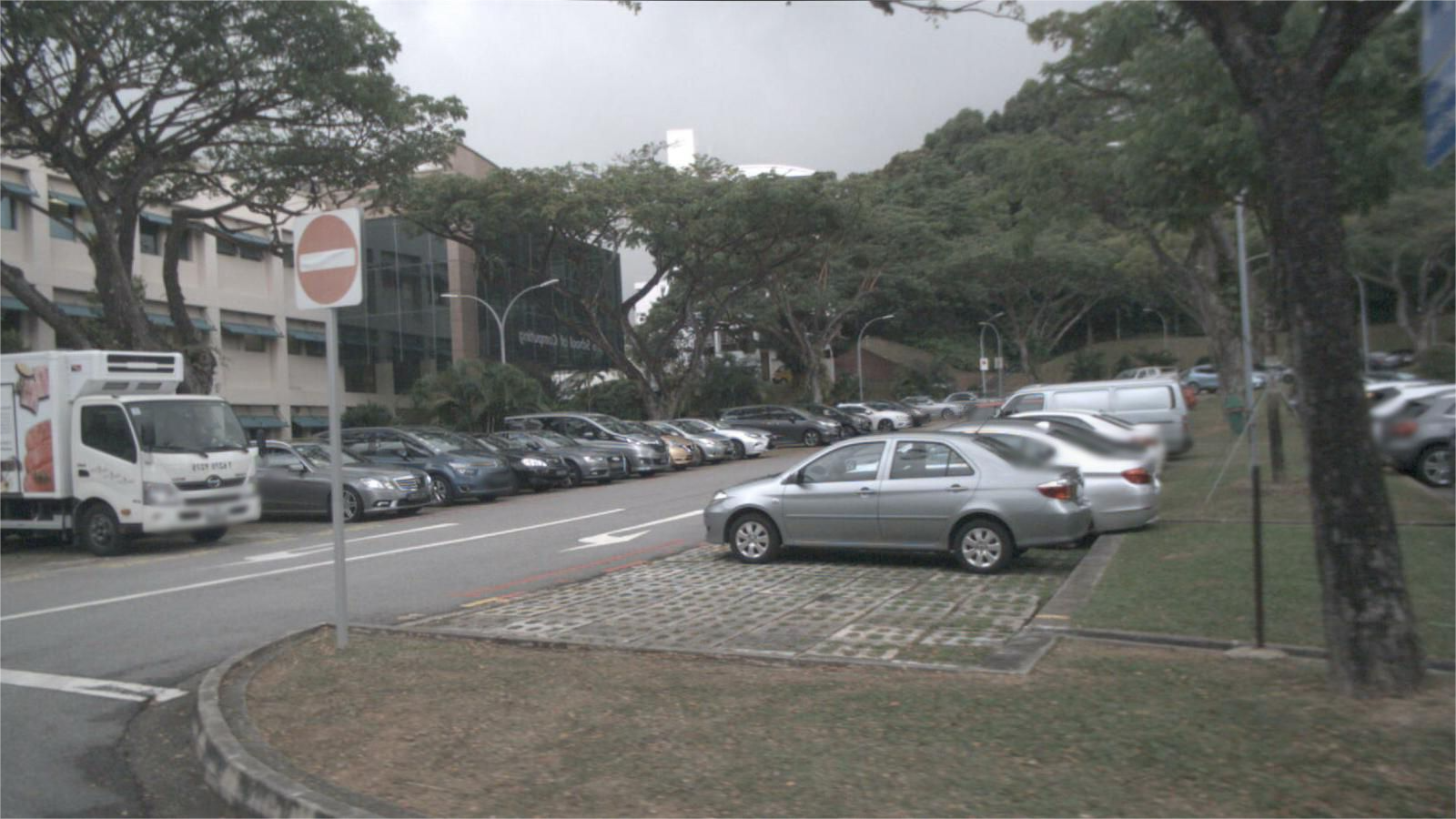}
        \caption{Real scene}
        \label{fig:sub.D2}
    \end{subfigure}
\caption{\label{fig:caseD}Scenario D, close on both sides at the corner}
\end{figure}

As shown in \autoref{fig:caseD}, involves the vehicle approaching from the bottom of the grid map, maneuvering around a bend, and then parking in a vertical parking spot in the parking lot.
The width of this parking spot is only about the width of a car, with both sides tightly bordered by road edges and other parked vehicles.

\subsubsection{Scenario E}
\begin{figure}[ht!]
  \centering
    \begin{subfigure}[t]{0.16\textwidth}
      \centering   
      \includegraphics[height=2.9cm,trim={0 0 25 25},clip]{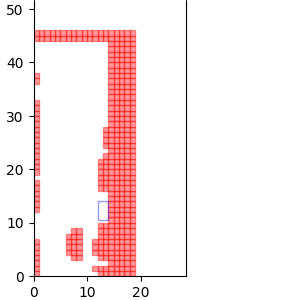}
        \caption{Grid map near the parking spot}
        \label{fig:sub.E1}
    \end{subfigure}
    \begin{subfigure}[t]{0.3\textwidth}
      \centering   
      \includegraphics[height=2.9cm]{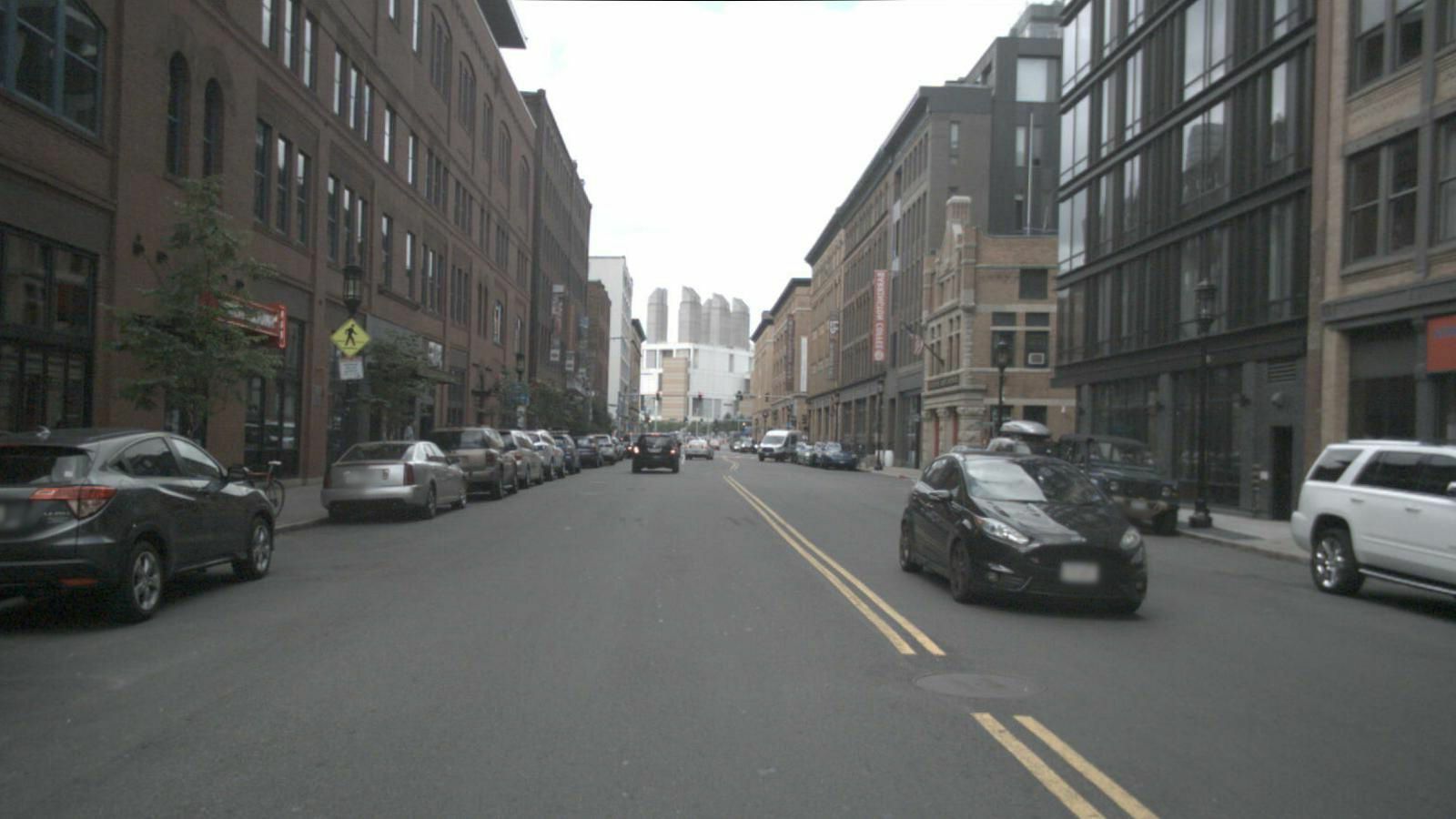}
        \caption{Real scene}
        \label{fig:sub.E2}
    \end{subfigure}
\caption{\label{fig:caseE}Scenario E, navigating around vehicles}
\end{figure}

Depicted in \autoref{fig:caseE}, requires the vehicle to approach the parking spot from below, navigate around a car, and then park in the right-side parking spot.
The vehicle must traverse a road where half the width is occupied by oncoming vehicles. 
The navigation algorithm needs to keep the vehicle as far away as possible from the vehicles on both sides to avoid collision.

\subsubsection{Scenario F}
\begin{figure}[ht!]
  \centering
    \begin{subfigure}[t]{0.17\textwidth}
      \centering   
      \includegraphics[height=2.9cm,trim={0 0 23 23},clip]{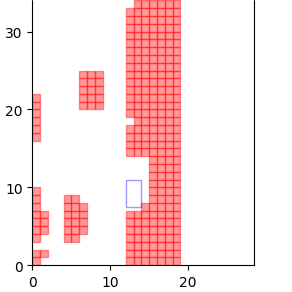}
        \caption{Grid map near the parking spot}
        \label{fig:sub.F1}
    \end{subfigure}
    \begin{subfigure}[t]{0.3\textwidth}
      \centering   
      \includegraphics[height=2.9cm]{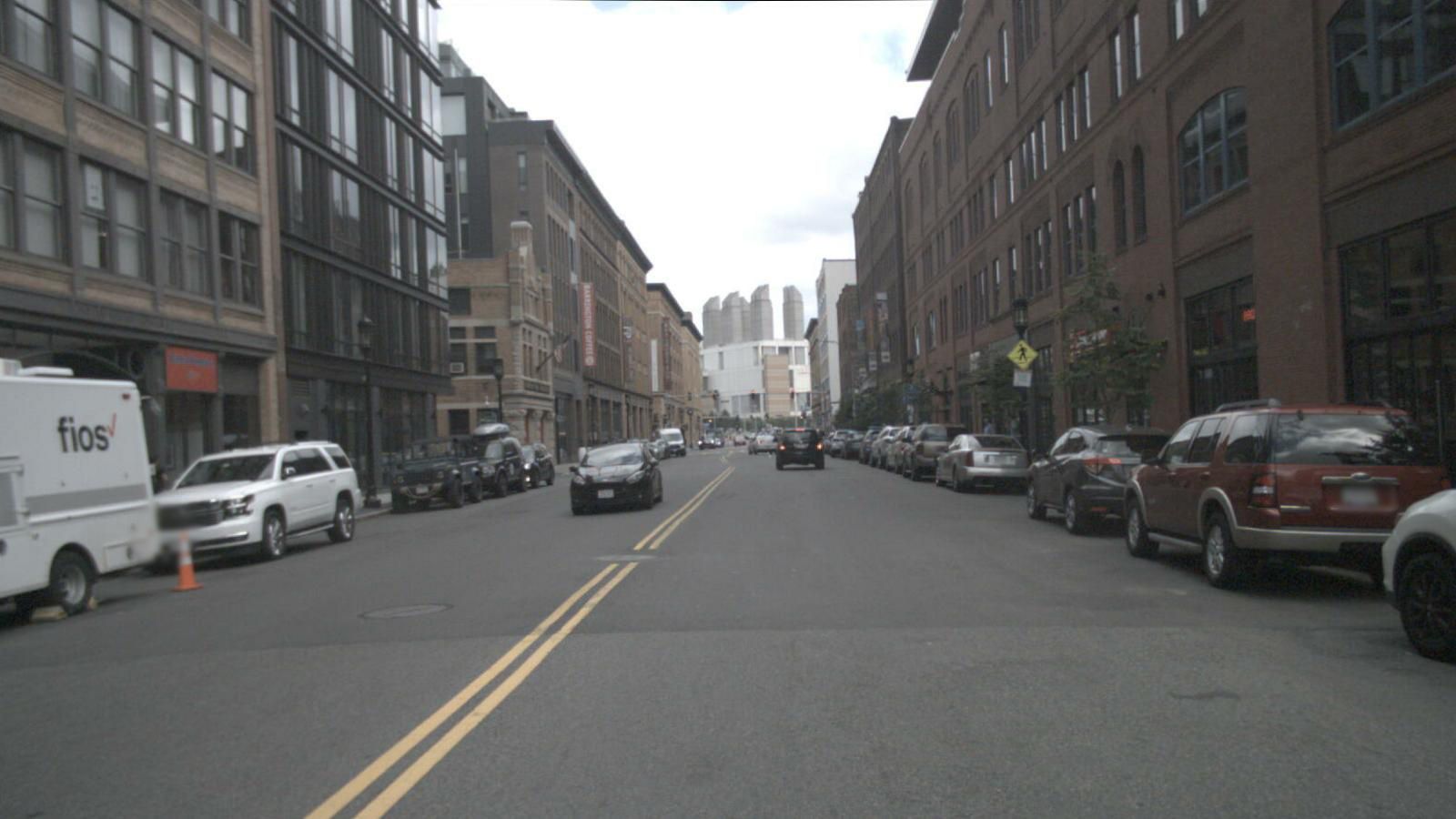}
        \caption{Real scene}
        \label{fig:sub.F2}
    \end{subfigure}
\caption{\label{fig:caseF}Scenario F, navigating around front and rear vehicles}
\end{figure}

As shown in \autoref{fig:caseF}, requires the vehicle to park in a parking spot to the front right. 
It is worth noting that both lanes of this road have passing vehicles, and the vehicle needs to navigate around these obstacles.

\subsection{Comfort Testing}
\label{sec:Comfort_Testing}

When driving a car, speed, acceleration, and steering frequency and magnitude are important indicators affecting driving comfort.
Literature indicates that higher speeds, higher acceleration and deceleration rates, and higher lateral accelerations result in lower passenger comfort \cite{ref50,ref51}.
Sudden changes in speed can lead to passenger discomfort, while excessive steering may cause passenger dizziness.
Therefore, this paper focuses on speed, acceleration, and steering angle to test the comfort of the proposed parking algorithm. Among them, acceleration and steering angle are the main indicators of concern.
The paper aims for a driving trajectory to have slow speed, low acceleration, small steering angles, and infrequent steering changes.

For parking trajectories, the paper additionally requires them to complete parking operations as quickly as possible, so parking time should be as short as possible while meeting the above requirements.

When testing the performance of different algorithms, this study uses the Model Predictive Control (MPC) algorithm to drive the car along the generated planning path in the simulation environment.
This study sets up penalty functions for automatic parking problems with MPC, which include acceleration and deceleration, steering, control command rate of change, vehicle position error, and parking pose error.
During testing, the same MPC controller is used to generate control commands, and the control command sequences corresponding to paths generated by different algorithms are compared to examine the speed, acceleration, and steering angle of the car during the driving process in the simulation testing environment.

\section{Results}
\label{sec:results}

\subsection{Algorithm Time Testing}
\label{sec:Algorithm_Time_Testing_res}

\subsubsection{Vertical Parking}
\label{sec:Vertical_Parking}

\begin{figure}[ht!]
  \centering
    \begin{subfigure}[t]{0.24\textwidth}
      \centering   
      \includegraphics[width=1\linewidth,trim={20 20 20 52},clip]{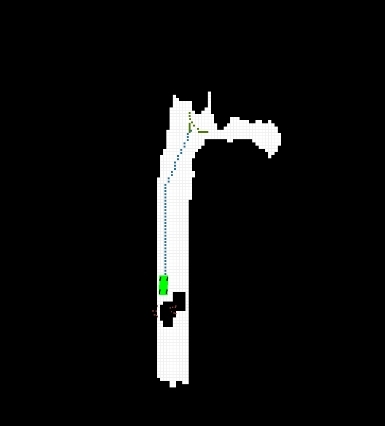}
        \caption{Basic A* algorithm planning result}
        \label{fig:A_V.sub1}
    \end{subfigure}
    \begin{subfigure}[t]{0.24\textwidth}
      \centering   
      \includegraphics[width=\linewidth,trim={20 20 20 50},clip]{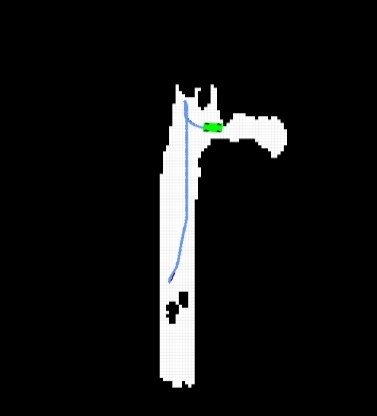}
        \caption{Improved A* algorithm planning result}
        \label{fig:A_V.sub2}
    \end{subfigure}
\caption{\label{fig:A_star_vertical}Comparison of Basic A* Algorithm and Improved A* Algorithm in Vertical Parking Scenario}
\end{figure}

This section discusses the vertical parking scenario, where the starting coordinates are (95, 85) and the target parking space coordinates are (109, 133).
The planning results of the two algorithms are shown in \autoref{fig:A_star_vertical}.
The original navigation path planning result image size is 200×200 pixels, the same as the size of the perception point cloud generated by SurroundOcc.
For ease of observation, some of the redundant black parts have been removed.

The planning result of the basic A* algorithm is shown in \autoref{fig:A_V.sub1}.
In the figure, the vehicle is represented in green; the drivable area in white; the non-drivable area in black, including but not limited to other vehicles, curbs, buildings, and pedestrians; the cyan path is generated by the basic A* algorithm from the starting point to the starting point of the parking path; the grass green path is from the parking start point to the final parking space.
This chapter focuses on the navigation paths generated by the basic A* algorithm and the improved A* algorithm. The simplified parking path generated by geometric curves is only for illustrative purposes.

It can be observed that the cyan path has an abrupt corner, making it impossible for a real vehicle to make a perfect sharp turn due to the minimum turning radius constraint.
Additionally, the transition between the cyan path and the grass green parking path is rather abrupt, and discontinuous paths will cause the vehicle to reposition its front wheels at each turn \cite{ref44}.
This is undoubtedly undesirable in this study.

The planning result of the improved A* algorithm proposed in this paper is shown in \autoref{fig:A_V.sub2}.
The path optimized after trajectory smoothing is shown in blue, and in the simulation environment, the vehicle also smoothly follows the path to the endpoint.
The optimized path no longer has abrupt turns and discontinuous curvature at the transitions, significantly improving the trajectory quality.

According to the metrics proposed in \autoref{sec:Algorithm_Time_Testing}, the recorded data is shown in \autoref{tab:comparison_vertical}:

\begin{table*}[!ht] \small \centering
\setlength\tabcolsep{2pt} \renewcommand{\arraystretch}{1.0}
    \caption{Algorithm Performance Comparison in Vertical Parking Scenario}
    \begin{tabular}{ccccccc}   \toprule
        Algorithm Time $(ms)$ & Initialization Time & Map Loading Time & Planning Time & Path Generation Time & Interpolation Time & Total Time  \\ \midrule
        Basic A* & 123.80 & 98255.68 & 21.86 & 0.02 & -- & 98401.36 \\ 
        Improved A* & 124.61 & 407.85 & 6.89 & 0.02 & 1.03 & 540.40 \\ \bottomrule
    \end{tabular} 
    \label{tab:comparison_vertical}
\end{table*}

From the table, it can be seen that the planning speed of the algorithm has significantly improved.
The optimized obstacle reading method and path planning method have greatly enhanced planning efficiency, reducing the total time for loading the map and planning the path from 98277.54 milliseconds to 413.072 milliseconds.
In the real map scenario of a local map, the planning speed has increased by more than 95\%.
In real-world map scenarios, the basic A* algorithm is almost incapable of planning an effective path within a limited time, whereas the improved A* can provide a path in a relatively short time, making it more valuable for practical use.

Thus, the efficiency of the improved A* algorithm has been greatly enhanced, making it suitable for complex automatic parking scenarios and high real-time requirements of local maps.

\subsubsection{Parallel Parking}
\label{sec:Parallel_Parking}

\begin{figure}[ht!]
  \centering
    \begin{subfigure}[t]{0.24\textwidth}
      \centering   
      \includegraphics[width=1\linewidth,trim={120 0 30 110},clip]{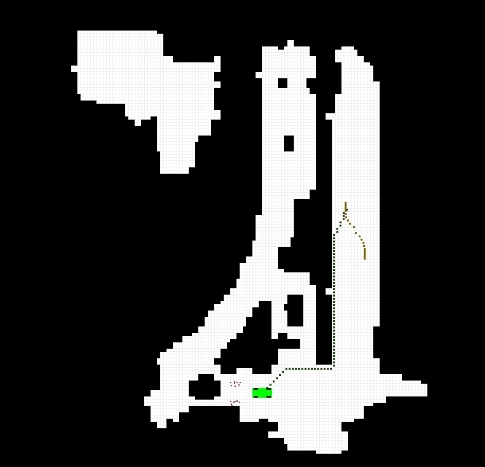}
        \caption{Basic A* algorithm planning result}
        \label{fig:A_P.sub1}
    \end{subfigure}
    \begin{subfigure}[t]{0.24\textwidth}
      \centering   
      \includegraphics[width=\linewidth,trim={120 0 30 110},clip]{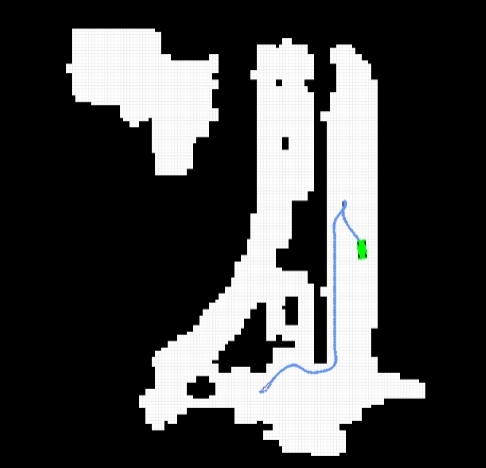}
        \caption{Improved A* algorithm planning result}
        \label{fig:A_P.sub2}
    \end{subfigure}
\caption{\label{fig:A_star_parallel}Comparison of Basic A* Algorithm and Improved A* Algorithm in Parallel Parking Scenario}
\end{figure}

The following map represents a parallel parking scenario, with the starting coordinates at (110, 65) and the target parking space coordinates at (142, 110).
The planning result of the basic A* algorithm is shown in \autoref{fig:A_P.sub1}. In the figure, the trajectory path is still represented by two different random colors to emphasize the seams in the path.
It should be noted that the traditional A* algorithm performs poorly in this scenario, with the planned path exhibiting severely unrealistic driving behaviors, such as overly close proximity to obstacles, which are highly dangerous.
Moreover, the planned trajectory requires the vehicle to make a right-angle turn and two acute-angle turns, with another acute-angle turn at the connection point of the parking path, ignoring the vehicle's physical performance constraints.

The planning result of the improved A* algorithm proposed in this paper is shown in \autoref{fig:A_P.sub2}.
It can be seen that the optimized blue trajectory has smoother turns and provides the vehicle with space to stay away from the wall.
The improved A* algorithm performs better than the basic A* algorithm in this scenario, making the trajectory more continuous and smooth while significantly reducing planning time.

The planning time data for the two algorithms in the parallel parking scenario are shown in \autoref{tab:comparison_parallel}.

\begin{table*}[!ht]  \small  \centering
\setlength\tabcolsep{1pt} \renewcommand{\arraystretch}{1.0}
    \caption{Algorithm Performance Comparison in Parallel Parking Scenario}
    \begin{tabular}{ccccccc} \toprule
        Algorithm Time $(ms)$ & Initialization Time & Map Loading Time & Planning Time & Path Generation Time & Interpolation Time & Total Time \\ \midrule
        Basic A* & 134.61 & 110479.79 & 65.77 & 0.03 & -- & 110680.20  \\ 
        Improved A* & 128.56 & 1196.75 & 20.32 & 0.03 & 1.469 & 1347.13 \\ \bottomrule
    \end{tabular}
    \label{tab:comparison_parallel}
\end{table*}

In this scenario, the improved A* algorithm again achieved an optimization rate of over 95\% in total time.
The initialization time for both the improved A* and basic A* remains nearly identical, but the map loading time and planning time are greatly reduced.
Furthermore, the traditional A* algorithm fails to plan a realistic path in this scenario, using a large number of abrupt, unrealistic turns and straight lines to form a path that a real vehicle could not follow.

This scenario further demonstrates that in realistic parking tasks considering the vehicle's dimensions, the improved A* algorithm proposed in this paper has a significant advantage in computation time, providing good real-time performance.

\subsubsection{Inaccessible Parking Spots}
\label{sec:Inaccessible_Parking_Spots_res}

This section tests the scenario where the parking spot is unreachable, with the starting coordinates at (95, 85) and the target coordinates at (109, 117).
It should be noted that the endpoint specified by the algorithm is an obstacle, making it impossible for the vehicle to reach the target point.
The endpoint is marked in red in the figure below.

\begin{figure}[ht!]  \centering
    \begin{subfigure}[t]{0.24\textwidth}
      \centering   
      \includegraphics[width=1\linewidth,trim={30 40 30 35},clip]{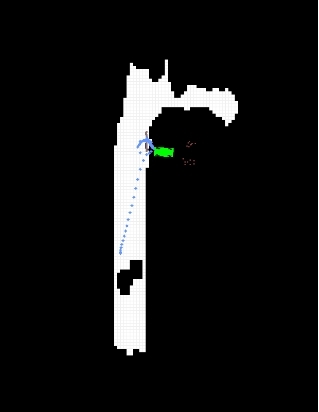}
        \caption{Basic A* algorithm unable to complete planning}
        \label{fig:unreach1}
    \end{subfigure}
    \begin{subfigure}[t]{0.24\textwidth}
      \centering   
      \includegraphics[width=\linewidth,trim={30 40 30 35},clip]{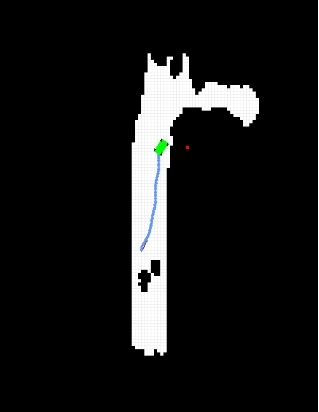}
        \caption{Improved A* algorithm still able to plan}
        \label{fig:unreach2}
    \end{subfigure}
\caption{\label{fig:unreach}Planning results of basic A* algorithm and improved A* algorithm when the parking spot is unreachable. The parking spot is red, located within the obstacle.}
\end{figure}

The basic A* algorithm cannot plan a path, leading to bizarre results from the controller.
Such results are clearly unusable in real hardware applications.
For demonstration purposes, the erroneous planning results are still shown in the simulation environment in this paper.
As shown in \autoref{fig:unreach1}, the path output by the planner (shown in purple) is evidently disconnected from the starting point, causing the trajectory executed by the controller (shown in blue) to be absurd and unrealistic.

The planning result of the improved A* algorithm proposed in this paper is shown in \autoref{fig:unreach2}.
Although the red point representing the parking endpoint is within the obstacle and unreachable, the improved A* algorithm still plans a smooth path as close to the endpoint as possible without being too close to the wall.
In this scenario, the initialization time is 121.57$(ms)$, the obstacle map loading time is 2.02$(ms)$, and the planning time is 326.19$(ms)$, maintaining sufficiently high performance.

The experiment demonstrates that the improved algorithm proposed in this paper can still complete the planning and navigate the vehicle as close to the target position as possible when the parking spot is unreachable.
The traditional A* algorithm does not have this capability.

\subsection{Dangerous Scenario Testing}
\label{sec:Dangerous_Scenario_Testing_res}

This paper uses two methods to test the scenarios proposed in \autoref{sec:Dangerous_Scenario_Testing}.
One method is the improved A* algorithm for planning the navigation path combined with a numerical optimization algorithm to generate the final parking path proposed in this paper.
The other method is the baseline A* algorithm combined with the geometric curve method as a control algorithm.

In the parking scenario tests described in this section, this paper compares the performance of the numerical optimization algorithm and the geometric curve method in generating parking paths in scenarios of varying complexity, examining whether both methods can generate feasible collision-free paths.
This paper will present detailed test results and a comparison table of the final parking path planning effects for certain scenarios.

\begin{figure*}[ht!]
  \centering
    \begin{subfigure}[t]{0.3\textwidth}
      \centering   
      \includegraphics[width=1\linewidth]{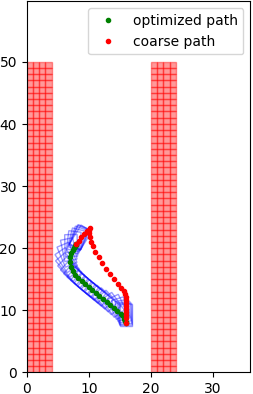}
        \caption{Planning results of both algorithms in Scenario B}
        \label{fig:B1}
    \end{subfigure}
    \begin{subfigure}[t]{0.63\textwidth}
      \centering   
      \includegraphics[width=\linewidth,trim={35 20 40 40},clip]{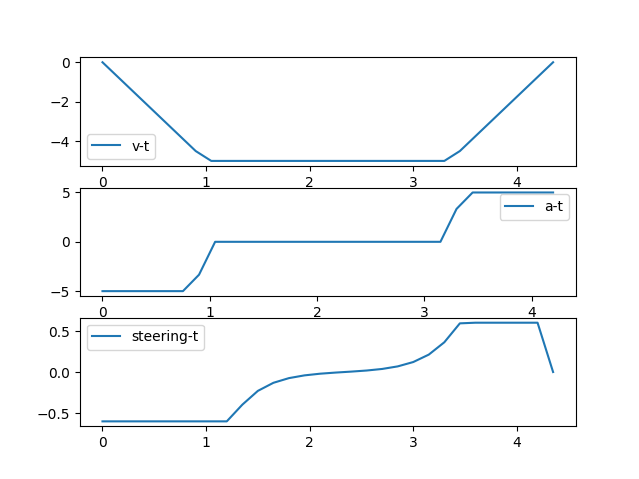}
        \caption{Driving parameters of the trajectory generated by the proposed parking algorithm in Scenario B}
        \label{fig:B2}
    \end{subfigure}
\caption{\label{fig:resultB}Comparison of planning results of both algorithms and driving parameters of the parking planning algorithm in Scenario B}
\end{figure*}

\autoref{fig:B1} shows the specific parking trajectories generated by the two algorithms.
The red trajectory is generated by the geometric curve method, while the green trajectory is generated by the numerical optimization algorithm proposed in this paper.
In test scenario B, both algorithms completed the parking maneuver, successfully reversing from the top left into the parking spot at the bottom right.

\autoref{fig:B2} shows the driving parameters of the parking trajectory generated by the numerical optimization method, with vehicle speed, acceleration, and steering wheel angle shown from top to bottom.
The data charts demonstrate that the numerical optimization method can generate continuous and stable driving trajectories, undoubtedly exhibiting high feasibility in control.

\begin{figure*}[ht!]
  \centering
    \begin{subfigure}[t]{0.4\textwidth}
      \centering   
      \includegraphics[width=1\linewidth,trim={0 1 0 0},clip]{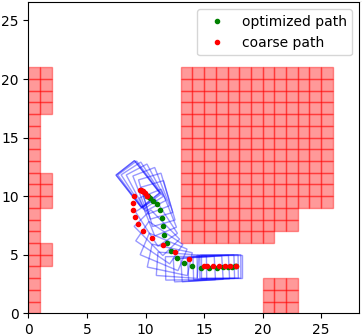}
        \caption{Planning results of both algorithms in Scenario C}
        \label{fig:C1}
    \end{subfigure}
    \begin{subfigure}[t]{0.5\textwidth}
      \centering   
      \includegraphics[width=\linewidth,trim={35 20 40 40},clip]{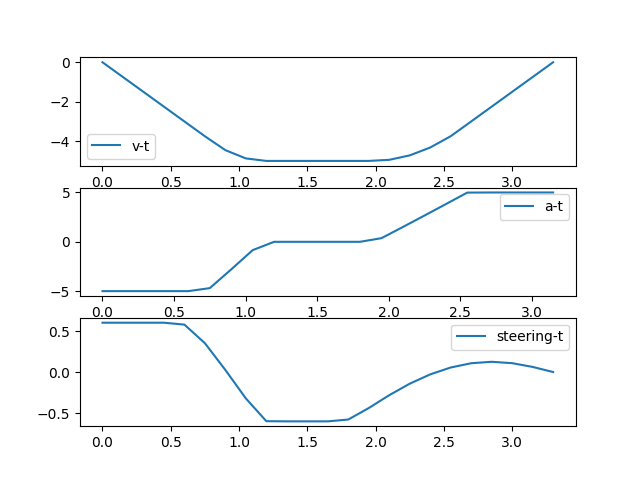}
        \caption{Driving parameters of the trajectory generated by the proposed parking algorithm in Scenario C}
        \label{fig:C2}
    \end{subfigure}
\caption{\label{fig:resultC}Comparison of planning results of both algorithms and driving parameters of the parking planning algorithm in Scenario C}
\end{figure*}

The parking trajectories generated by the two algorithms in Scenario C are shown in \autoref{fig:C1}.
Both algorithms generated a reversing path into the bottom right vertical parking spot.
It can be observed that when facing a parking spot close to other obstacles, the numerical optimization method can fully utilize the space to generate a safe path.
In contrast, the geometric curve method tends to generate paths that are too close to obstacles, posing a high risk of collision.

By examining the driving data of the parking trajectory generated by the numerical optimization method shown in \autoref{fig:C2}, it can be found that even in situations with limited space on one side of the parking spot, the numerical optimization method can still ensure stable driving.

\begin{figure*}[ht!]
  \centering
    \begin{subfigure}[t]{0.23\textwidth}
      \centering   
      \includegraphics[width=1\linewidth]{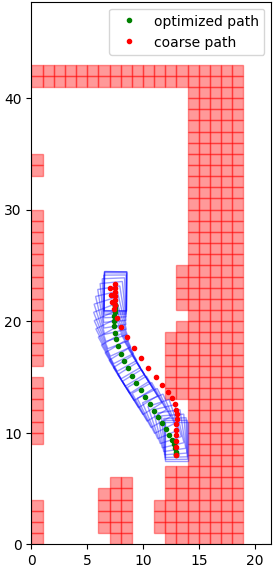}
        \caption{Planning results of both algorithms in Scenario E}
        \label{fig:E1}
    \end{subfigure}
    \begin{subfigure}[t]{0.66\textwidth}
      \centering   
      \includegraphics[width=\linewidth,trim={25 160 40 40},clip]{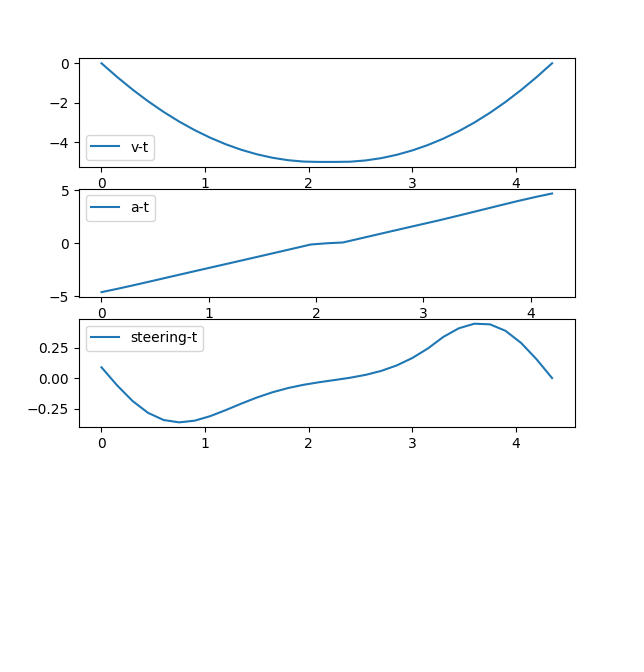}
        \caption{Driving parameters of the trajectory generated by the proposed parking algorithm in Scenario E}
        \label{fig:E2}
    \end{subfigure}
\caption{\label{fig:resultE}Comparison of planning results of both algorithms and driving parameters of the parking planning algorithm in Scenario E}
\end{figure*}

As shown in \autoref{fig:E1}, in Scenario E, due to the limited parking space, the geometric curve method cannot generate a feasible path, with the resulting path (red path) significantly encroaching into the space occupied by obstacles. 
In contrast, the numerical optimization method used in this paper can achieve parking in narrow spaces.
As shown in \autoref{fig:E2}, the parking operation is accomplished with a single attempt to reverse into the spot, without repeated unnecessary maneuvers, resulting in a clean and efficient path (green path).

The comparison results of the hazardous scenario tests are shown in \autoref{tab:danger result}.
In all test scenarios, the geometric curve method only succeeded in the two simplest test scenarios, Scenario A and Scenario B.
In Scenario C, the path planned by the geometric curve method caused the vehicle to scrape against the boundary.
In the remaining three scenarios, due to the narrow parking space and numerous surrounding obstacles, the geometric curve method failed to plan a collision-free path.

On the other hand, the numerical optimization algorithm proposed in this paper successfully planned feasible parking paths in all six test scenarios, fully utilizing the narrow space to complete the parking task.
The comparison indicates that the numerical optimization method used in this paper can pass more test cases and, compared to the commonly used geometric curve method, generates parking trajectories with higher feasibility and safety.

\begin{table}[!ht] \small \centering
\setlength\tabcolsep{2pt} \renewcommand{\arraystretch}{1.0}
\caption{Comparison of hazardous scenario test results between the optimized algorithm proposed in this paper and the baseline algorithm}
\begin{tabular}{lcccccc} \toprule
Scenario & A & B & C & D & E & F \\ \midrule
Traditional Geometric Curve Method & \checkmark & \checkmark & \ding{53} & \ding{53} & \ding{53} & \ding{53} \\
Numerical Optimization Method in This Paper & \checkmark & \checkmark & \checkmark & \checkmark & \checkmark & \checkmark \\ \bottomrule
\end{tabular}
\label{tab:danger result}
\end{table}

\subsection{Comfort Testing}
\label{sec:Comfort_Testing_res}

\begin{figure}[!ht] \centering
\includegraphics[width=0.98\linewidth,trim={35 12 55 48},clip]{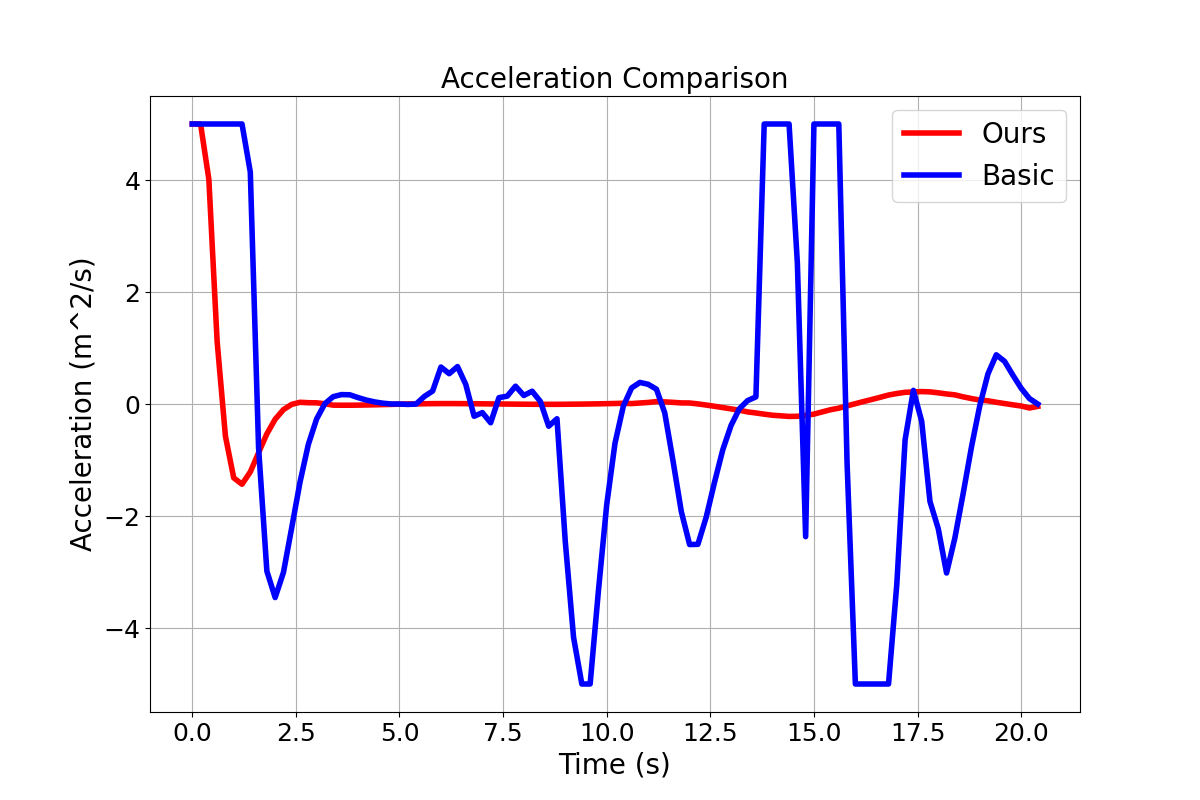}
\caption{\label{fig:acc}Comparison of Accelerations for Navigation Paths Generated by Baseline A* and Improved A* Algorithms}
\end{figure}

\begin{figure}[!ht] \centering
    \includegraphics[width=0.96\linewidth,trim={35 12 55 48},clip]{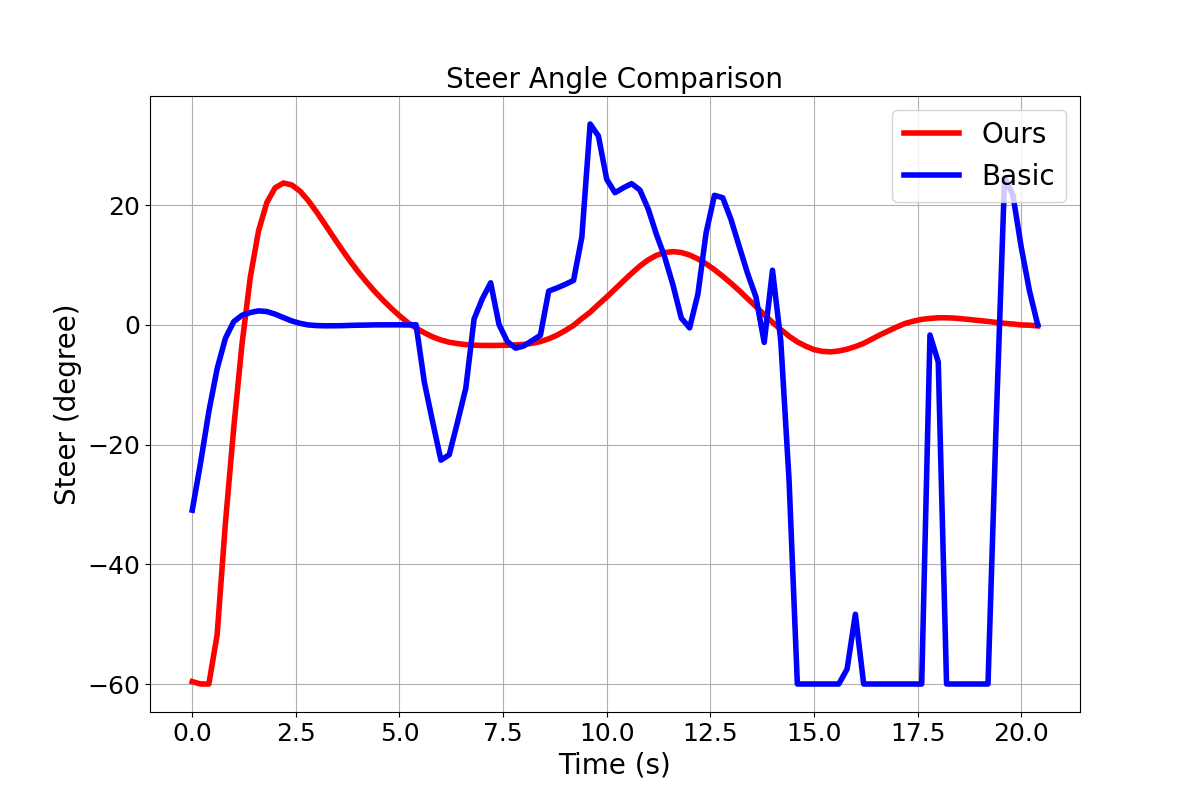}
    \caption{\label{fig:steer}Comparison of Steering Angles for Navigation Paths Generated by Baseline A* and Optimized A* Algorithms.}
\end{figure}

In \autoref{sec:Vertical_Parking} and \autoref{sec:Parallel_Parking}, the proposed improved A* algorithm is compared with the baseline A* algorithm. 
This section provides a graphical comparison of the comfort of the trajectories generated by these two methods. 
The driving parameter statistics for the two test scenarios are shown in \autoref{tab:data_vertical} and \autoref{tab:data_parallel}, and the acceleration and steering angle curves are shown in \autoref{fig:acc} and \autoref{fig:steer}.

\begin{table}[!ht] \small \centering
\setlength\tabcolsep{2pt} \renewcommand{\arraystretch}{1.0}
\caption{Driving Parameters for the Baseline A* and Improved A* Algorithms in the Vertical Parking Scenario}
\begin{tabular}{ccc} \toprule
Algorithm & \makecell{Average \\ Acceleration $(m/s^{2})$} & \makecell{Average Steering \\ Angle $(\degree)$} \\ \midrule
Baseline A* & 1.81 & 19.74 \\
Improved A* & 0.30 & 10.79 \\ \bottomrule
\end{tabular}
\label{tab:data_vertical}
\end{table}

The data in \autoref{tab:data_vertical} indicate that in the vertical parking scenario, the average acceleration of the trajectory planned by the baseline A* is 1.81 ($m/s^2$), while the average acceleration of the trajectory planned by the improved A* is only 0.30 ($m/s^2$), representing an optimization of 83.4\%. 
The average steering angle for the baseline A* trajectory is 19.74°, while for the improved A* trajectory, it is only 10.79°, representing an optimization of 45.3\%.

\begin{table}[!ht] \small \centering
\setlength\tabcolsep{2pt} \renewcommand{\arraystretch}{1.0}
\caption{Driving Parameters for the Baseline A* and Improved A* Algorithms in the Parallel Parking Scenario}
\begin{tabular}{cccc} \toprule
Algorithm & \makecell{Average \\\ Acceleration $(m/s^{2})$} & \makecell{Average Steering \\ Angle (\degree)} \\ \midrule
Baseline A* & 1.49 & 12.70 \\
Improved A* & 0.30 & 20.03 \\ \bottomrule
\end{tabular}
\label{tab:data_parallel}
\end{table}

\autoref{tab:data_parallel} shows that the improved A* algorithm has an optimization of 79.8\% in terms of acceleration. 
However, the average steering angle data in this scenario is not meaningful, as the baseline A* algorithm's planned path presented in \autoref{sec:Parallel_Parking} cannot be used for actual driving tasks.

From \autoref{tab:data_vertical} and \autoref{tab:data_parallel}, it is evident that in both test scenarios, the vehicle experiences smaller accelerations and smaller average steering angles while following the navigation path.
This indicates that acceleration and deceleration, as well as steering, are smoother, reducing unnecessary control and improving the ride comfort for passengers.

In the vertical parking scenario of \autoref{sec:Vertical_Parking}, the acceleration and steering angle curves for the vehicle's driving process using both algorithms are shown in \autoref{fig:acc} and \autoref{fig:steer}.

\autoref{fig:acc} compares the accelerations corresponding to the baseline A* algorithm and the optimized A* algorithm. 
The horizontal axis represents the simulation vehicle's driving time, and the vertical axis represents the acceleration. 
The sampling range is the navigation path from the simulation vehicle's starting point to the starting point of the parking path. 
The blue curve represents the acceleration curve corresponding to the baseline A* algorithm, and the red curve represents the acceleration curve corresponding to the proposed improved A* algorithm. 
It can be observed that the acceleration curve corresponding to the improved algorithm is very smooth, while the acceleration curve corresponding to the baseline A* algorithm is very unstable. 
This study suggests that the baseline A* algorithm generates a planning path with many sharp angles and curvature discontinuities, causing the controller to repeatedly accelerate and decelerate to adapt to the path.
However, the optimized A* algorithm path is smoother and more continuous, so the controller does not need to repeatedly adjust the vehicle's speed.

\autoref{fig:acc} shows that compared to the baseline A* method, the proposed improved A* method significantly enhances real-time performance while improving trajectory quality.

\autoref{fig:steer} compares the steering angles corresponding to the baseline A* algorithm and the optimized A* algorithm. 
Similarly, it can be seen that the path generated by the baseline A* algorithm forces the planner to adopt a violent driving behavior with repeated full steering lock, while the path generated by the improved A* algorithm allows the planner to use smoother operations to control the vehicle.

\autoref{tab:data_vertical} and \autoref{tab:data_parallel}, along with \autoref{fig:acc} and \autoref{fig:steer}, show that the improved A* algorithm generates higher quality planning trajectories.
The paths are shorter, unnecessary accelerations and steering adjustments are greatly reduced, and the vehicle's driving process is safer, more stable, and more comfortable.

\begin{figure}[ht!]
  \centering
    \begin{subfigure}[t]{0.495\textwidth}
      \centering   
      \includegraphics[width=1\linewidth,trim={35 20 40 41},clip]{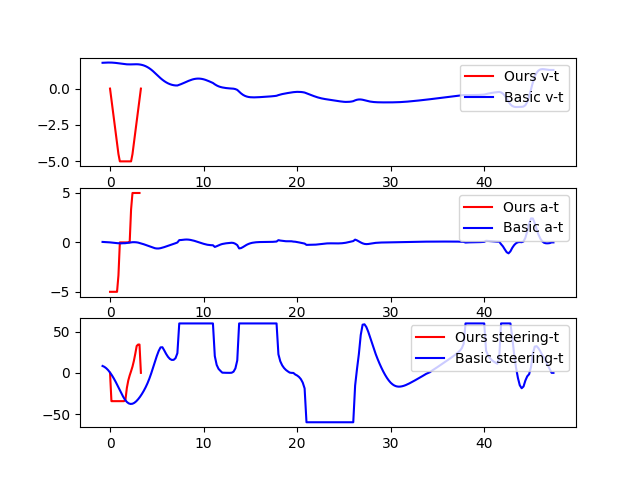}
        \caption{Scenario B}
        \label{Fig.sub.1}
    \end{subfigure}
    \begin{subfigure}[t]{0.495\textwidth}
      \centering   
      \includegraphics[width=\linewidth,trim={35 20 40 41},clip]{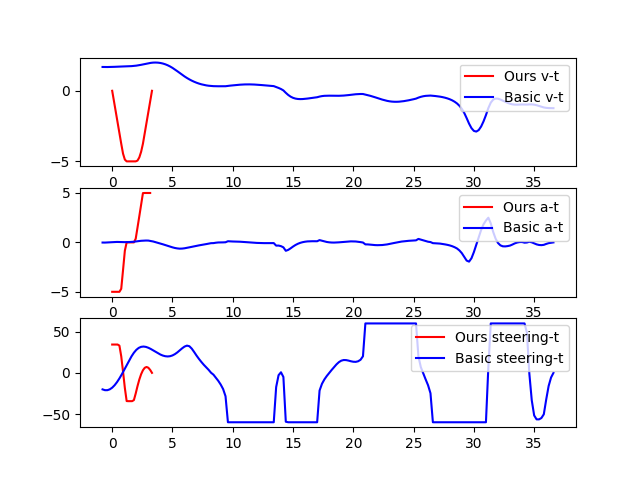}
        \caption{Scenario C}
        \label{Fig.sub.2}
    \end{subfigure}
\caption{\label{fig:data_com}Comparison of Speed, Acceleration, and Steering Angle for Parking Trajectories Generated by the Proposed Method and Traditional Method in Some Hazardous Test Cases.}
\end{figure}

Next, this study compares the comfort of parking paths generated by the numerical optimization-based parking algorithm and the common geometric curve method.
In \autoref{fig:data_com}, the red curve represents the data corresponding to the proposed algorithm, and the blue curve represents the data corresponding to the geometric curve method. 
It is apparent that the trajectory corresponding to the geometric curve method appears hesitant and repetitive, with very frequent and large steering adjustments. 
The trajectory output by the numerical optimization method, while having greater acceleration, allows the vehicle to complete parking with decisive maneuvers, greatly saving the driver's and passengers' time.

Combining the above content, it can be seen that the proposed automatic parking planning scheme, which generates navigation paths using the improved A* algorithm and then outputs parking trajectories via the numerical optimization method, has comfort advantages throughout the process compared to the traditional and widely used A* combined with geometric curve method.

\section{Discussion}
\label{sec:discussion}

The experiments in this paper consist of two parts: algorithm time tests designed for the improved A* algorithm used for generating navigation paths, and safety and comfort tests designed for the numerical optimization method used for generating parking trajectories.

Through a series of experiments in different scenarios, this paper compares and contrasts the time required by the improved A* algorithm and the traditional A* algorithm to generate navigation paths.
The results show that, considering the volume of the ego vehicle, the proposed improved A* algorithm generally achieved a computational speed increase of 95\% or more, with particularly significant improvements in map loading time and planning time.
This demonstrates that the series of performance optimization methods designed in this paper are feasible and effective, addressing issues such as excessive search nodes and prolonged single search times in the traditional A* algorithm, thereby significantly enhancing the planning time efficiency of the A* algorithm.

In the dangerous test case experiments, this paper compares the proposed numerical optimization-based parking planning method with the traditional geometric curve method to test the ability of the proposed parking method to complete parking trajectory planning tasks in various scenarios. 
The experimental results show that the proposed numerical optimization method can accomplish parking planning in scenarios ranging from simple to difficult, and the number of test cases passed in dangerous scenarios is 200\% higher than that of the geometric curve method, indicating higher safety in the generated trajectories.

In the comfort validation, this paper comprehensively tests the overall planning process proposed, collects vehicle motion parameters returned by the controller, and compares them with the traditional method combining the basic A* with the geometric curve method.
The data show that the paths generated by the proposed method result in smooth vehicle movement without frequent speed and direction changes, leading to more decisive vehicle behavior.

The proposed method in this paper demonstrates better performance in real-time and safety metrics, which are crucial for autonomous parking tasks, and also offers higher comfort compared to traditional methods.

\subsection{Limitations}
\label{sec:Limitations}

Shortcomings and potential improvements for the proposed method are evident. 
For example, when there are many obstacles near the parking space or the parking space is too narrow, the computational time of the numerical optimization algorithm significantly increases.
The real-world situation may not fully meet all the assumptions of the model in \autoref{sec:Vehicle_Model}, and this paper has not yet considered dynamic constraints such as tire slip, which can have a significant impact on vehicle motion.

In future work, this paper will further improve the algorithm and attempt to combine it with other advanced technologies to further enhance the real-time performance and safety of autonomous parking planning.

\section{Conclusions}
\label{sec:conclusion}

In this paper, the characteristics of current automated parking tasks are summarized, traditional perception methods are discussed, and the advantages and disadvantages of conventional automated parking planning methods are analyzed.
The study reveals that the automated parking task under BEV local maps poses higher demands on the real-time and safety aspects of planning algorithms, which traditional methods cannot meet.
Therefore, this paper proposes a novel automated driving parking planning scheme suitable for BEV local maps.

This paper successfully achieves superior planning results by targeting improvements and linking traditional methods.
The main advantages of the proposed approach include:
\begin{enumerate}
    \item Using an improved A* algorithm to generate navigation paths, enhancing algorithm computational speed to ensure real-time performance.
    \item Employing numerical optimization methods to generate parking paths, ensuring safety.
    \item Providing a higher level of comfort in path generation compared to traditional methods.
\end{enumerate}

The proposed algorithm is tested in a Python testing environment and simulation simulator alongside traditional automated parking planning methods. 
Tests include algorithm runtime, dangerous scenario use cases, and comfort validation.
Experimental results demonstrate significant improvements in real-time performance, safety, and comfort compared to traditional methods.
This approach is better suited for addressing the requirements of automated parking planning tasks under BEV local maps.
In the future, the knowledge graph \cite{liu2024mining,lan2022semantic,liu2022towards} and machine learning methods \cite{gao2021neat,lan2022class} could be used to improve this work.

\bibliographystyle{IEEEtran}
\bibliography{bibliography}

\end{document}